\documentclass[default,iicol]{sn-jnl}

\pdfoutput=1 
\usepackage{booktabs}

\usepackage{graphicx}%

\usepackage{amsmath,amssymb,amsfonts}%
\usepackage{makecell}

\usepackage{float}

\usepackage[T1]{fontenc}
\usepackage{lmodern}
\usepackage{microtype}

\usepackage{amsthm}%
\usepackage{mathrsfs}%
\usepackage[title]{appendix}%
\usepackage{xcolor}%
\usepackage{textcomp}%
\usepackage{manyfoot}%
\usepackage{booktabs}%
\usepackage{algorithm}%

\usepackage{algpseudocode}%
\usepackage{listings}%
\usepackage{subfigure}

\usepackage{textcomp}

\usepackage[dvipsnames]{xcolor}
\usepackage{colortbl}
\usepackage{array}
\usepackage{xcolor}

\usepackage{multirow}

\usepackage{arydshln}
\usepackage{multicol}

\usepackage{color}

\usepackage{geometry}
\usepackage{colortbl}

\definecolor{tabfirst}{rgb}{1, 0.75, 0.7}
\definecolor{tabsecond}{rgb}{1, 0.83, 0.7}
\definecolor{tabthird}{rgb}{1, 0.96, 0.7}

\usepackage[numbers]{natbib}
\begin{document}

\title[Article Title]{Voxelized 3D Gaussian Representation for Dense Visual SLAM on Embedded Vision System}


\author[1]{\fnm{Tianchen} \sur{Deng}}

\author[1]{\fnm{Chang} \sur{Nie}}

\author[3]{\fnm{Shuhong} \sur{Liu}}

\author[1]{\fnm{Wenhua} \sur{Wu}}

\author[2,4]{\fnm{Jianfei} \sur{Yang}}

\author[2]{\fnm{Shenghai} \sur{Yuan}}

\author[5]{\fnm{Jiuming} \sur{Liu}}

\author[1]{\fnm{Zhe} \sur{Liu}}

\author[2]{\fnm{Danwei} \sur{Wang}}

\author[1]{\fnm{Hesheng} \sur{Wang}}

\affil[1]{\orgname{Shanghai Jiao Tong University}}
\affil[2]{\orgname{Nanyang Technological University}}
\affil[3]{\orgname{The University of Tokyo}}
\affil[4]{\orgname{Harvard University}}
\affil[5]{\orgname{University of Cambridge}
}



\abstract{
Recent work has shown that 3D Gaussian-based SLAM enables high-quality reconstruction, accurate pose estimation, and real-time rendering of scenes. However, when we revisit the existing GS-based SLAM systems, we observe that the redundancy of the Gaussian ellipsoids created by SLAM systems is significantly superior to those of the original 3D Gaussian Splatting methods. We further quantitatively demonstrate the geometric similarity and the redundancy of 3D Gaussian scene representation, leading to high memory and storage costs and limiting its broader application in embedded platforms. To address this limitation, we propose a voxelized compact 3D Gaussian representation for SLAM system that reduces the number and the parameter size of Gaussian ellipsoids. A novel voxel-anchored 3D Gaussian scene representation method is proposed to achieve compact and accurate scene reconstruction. A sliding window-based online masking strategy is first proposed to reduce the redundant ellipsoids. Then, a novel residual codebook-based quantization method is proposed to further compress 3D Gaussian attributes. Robust and accurate pose estimation is achieved by a local-to-global bundle adjustment method with icp loss. Extensive experiments demonstrate that our method achieves faster training, rendering speed (\textbf{226$\%$} increase), and low memory usage (\textbf{2.21$\times$} compression) while maintaining the state-of-the-art (SOTA) quality of the scene representation. Moreover, we also test our framework on an embedded platform, Jetson Xavier MIC-730AI, showing the potential of application in embedded system. We also collect a Neural SLAM dataset using our mobile robot's embedded hardware platform, incorporating data from multiple sensor modalities to further validate the effectiveness of our algorithm on a real-world system across various indoor and outdoor scenes. We will release our code  in \href{https://github.com/dtc111111/VCGS-SLAM}{https://github.com/dtc111111/VCGS-SLAM}
}

\keywords{Embedded Vision, 3D Gaussian Splatting, Compact Dense Visual SLAM, Model Compression and Acceleration}



\maketitle
Simultaneous localization and mapping (SLAM) 
has been a fundamental embedded vision problem with wide applications such as autonomous driving, robotics, and virtual/augmented reality~\cite{slam,deng2025best3dscenerepresentation}. 
Several traditional methods, including ORBSLAM~\cite{orbslam}, and~\cite{xie,past,wz2}, have been introduced over the years, representing scenes with sparse point cloud maps. However, sparse map representations fail to provide sufficient scene information and cannot meet the requirements of robotic downstream tasks such as navigation, path planning, and obstacle avoidance. Attention has turned to dense scene reconstruction, exemplified by DTAM~\cite{dtam}, Kintinuous~\cite{Kintinuous}, and ElasticFusion~\cite{elasticfusion}. However, their accuracy remains unsatisfactory due to high memory costs, slow processing speeds, and real-time running limitations.

\begin{figure*}[t]
  \centering
\includegraphics[width=\linewidth]{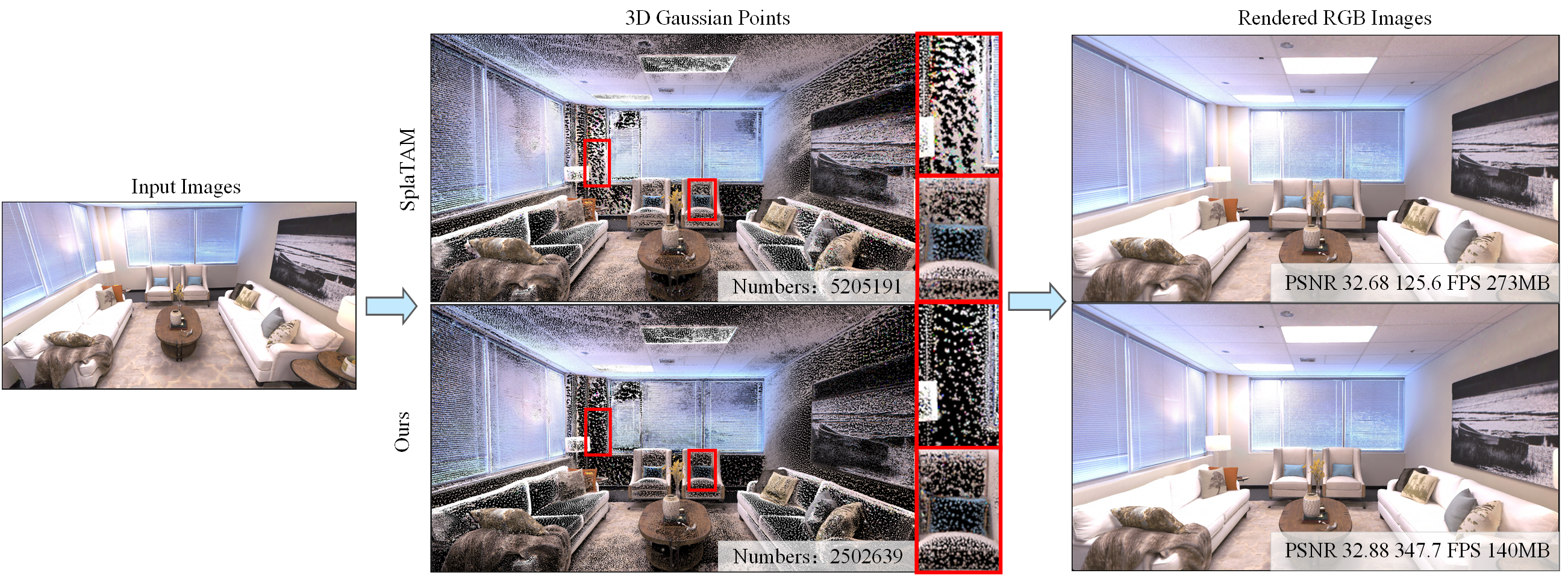}
  \caption{Our framework minimizes storage and accelerates rendering while maintaining the SOTA image reconstruction performance. The proposed framework eliminates unnecessary 3D Gaussian ellipsoids without affecting performance. We highlight and enlarge some areas to show the significant reduction of 3D Gaussian points.}
  \label{fig:first}
\end{figure*}

Nowadays, with the proposal of Neural Radiance Fields
(NeRF), many works focus on combining implicit scene representation with SLAM systems. iMAP~\cite{imap} is the first method to use a single MLP to represent the scene. NICE-SLAM~\cite{niceslam}, ESLAM~\cite{eslam}, Co-SLAM~\cite{coslam}, and so on~\cite{wz1} further improve the scene representation with the hybrid feature grids, axis-aligned feature planes, joint coordinate-parametric encoding, and progressive scene representation. 
Recent methods have started to explore 3D Gaussian Splatting(GS)~\cite{3dgs} to further improve the speed of rendering~\cite{fmgs}. SplaTAM~\cite{splatam}, GS-SLAM~\cite{gsslam}, Mono-GS~\cite{monogs}, Gaussian-SLAM~\cite{gaussianslam} are the pioneer works. GS-based SLAM methods 
leverage a point-based representation associated with 3D Gaussian attributes and adopt the rasterization pipeline to
render the images, achieving fast rendering speed and promising image quality. However, we revisit the performance of various GS-based SLAM systems~\cite{splatam,gsslam,monogs,gaussianslam} and perform a quantitative analysis and validation of the geometric similarity. We have noticed that the Gaussian ellipsoids generated by SLAM systems exhibit significantly greater geometric similarities compared to those produced by the original 3D Gaussian Splatting method.
This similarity stem from the optimization approach of the SLAM system. They usually need more than 500MB to represent a small room-sized scene, which hinders practical deployment, especially on resource-constrained embedded devices.

To address the scene representation redundancy, we propose a compact 3D Gaussian-based SLAM framework to address the critical high memory demand and slow training speed issue in GS-based SLAM systems. Our method notably enhances storage efficiency while delivering high-quality reconstruction, fast training speed, and real-time rendering capabilities. First, we design a voxel-anchored sequential 3D Gaussian representation method for compact SLAM framework. A novel sliding window-based online masking method tailored for SLAM framework is proposed to remove the millions of redundant and unnecessary 3D Gaussian ellipsoids created during the SLAM system operation, achieving faster rendering speed and efficient memory usage since the computational complexity is linearly proportional to the number of 3D Gaussian ellipsoids.

Second, a codebook-based quantization method is designed to compress the attributes of each Gaussian ellipsoid and neural anchor point. It learns to find the similarities shared across the scene. We only store the codebook index for each 3D anchor point, obtaining compact scene representation.

Third, the camera tracking accuracy of GS-based SLAM is relatively low compared with other SLAM systems. A local-to-global BA method with icp loss is proposed to achieve robust and accurate pose estimation. Our method maintains a sparse global keyframe database and performs bundle adjustment with all the historical observations, which can effectively eliminate cumulative error. \textbf{Overall, our contributions are shown as follows:}
 \begin{itemize}
    \item We revisit the existing GS-based SLAM systems of the scene representation. We quantitatively analyze and demonstrate this geometric similarity, theoretically proving the feasibility and significance of compressing the scene representation in GS-based SLAM systems.
    \item We propose a novel compact 3D Gaussian SLAM framework, achieving fast training and rendering speed, accurate pose estimation, and significantly enhancing storage efficiency.
    \item We design a voxel-anchored 3D Gaussian representation for compact and accurate scene reconstruction. A novel sliding window-based online masking method is proposed to remove the number of redundant Gaussian ellipsoids while achieving high-fidelity performance during training. A Kmeans-based codebook quantization method to efficiently restore the information of each neural anchor point during the SLAM system operation. A keyframe-based local-to-global BA method with icp loss is proposed to improve the relatively low performance of camera tracking. 
    \item We conduct comprehensive experiments on different datasets and embedded platforms, such as Jetson and laptop. We achieve nearly 226\% increase in rendering speed and over 2.21$\times$ compression on memory usage.
\end{itemize}

\begin{figure*}[t]
  \centering
\includegraphics[width=\linewidth]{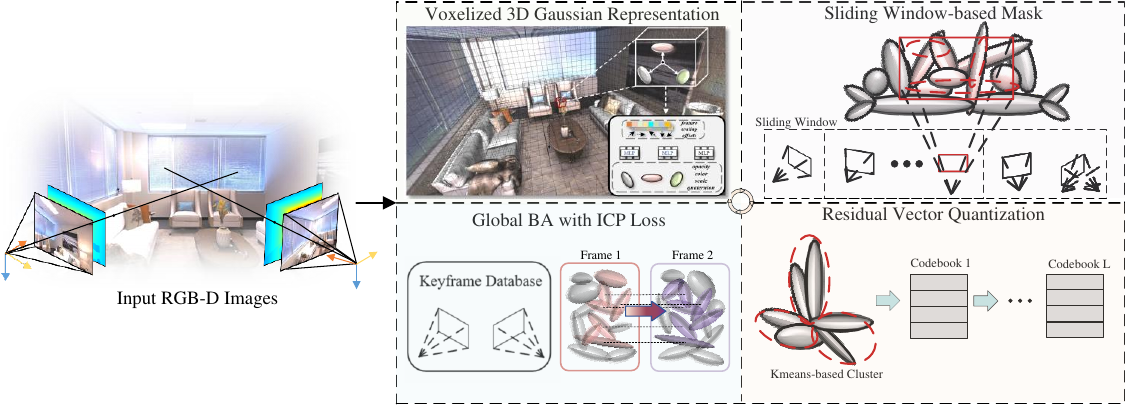}
  \caption{The pipeline of our GS-based SLAM system. The input of our system is RGB-D images. We start the SLAM system by initializing the 3D Gaussian map construct. Then, we update our 3D Gaussian map by adding new Gaussians and using the learnable mask to reduce the redundant 3D Gaussian ellipsoids. We incorporate a codebook-based vector quantization method to compress the scene representation. For camera tracking, we maintain a global keyframe database for local-to-global BA and use icp loss for robust pose estimation. }
  \label{fig:pipeline}
  
\end{figure*}
\section{Related Work}
\textbf{Dense Visual SLAM.} SLAM has become an active field for the past two decades~\cite{slam2,hscnet++,slam3}. Traditional visual SLAM algorithms~\cite{orbslam} estimate accurate camera poses and use sparse point
clouds as the map representation. They use manipulated key points for tracking, mapping, relocalization, and loop closing. DTAM~\cite{dtam} is the first method to achieve dense scene reconstruction. Kinectfusion~\cite{kinectfusion} uses projective iterative-closet-point (ICP) for camera tracking. Some learning-based methods integrate traditional geometry frameworks with deep learning networks for accurate camera tracking and mapping, such as DROID-SLAM~\cite{droidslam}. It achieves impressive trajectory estimations by
using neural networks to leverage richer context from images with RAFT~\cite{raft} feature. CodeSLAM~\cite{codeslam} propose a dense representation of scene geometry which is conditioned on the intensity data
from a single image and generated from a code consisting of a small number of parameters.   
DPVO~\cite{dpvo} propose a novel recurrent network architecture designed
for tracking image patches across time.  In contrast to
previous SLAM approaches, we adopt implicit scene representation of the geometry and directly optimize them during mapping for accurate and dense scene representation.

\noindent\textbf{NeRF-based SLAM.} There are many scene representation methods~\cite{pseudo} for neural SLAM system such as point cloud, voxel and so on. With the proposal of
Neural radiance fields (NeRF)~\cite{NeRF}, many researchers explore taking advantage of the implicit method into
SLAM systems. 
iMAP~\cite{imap} is the first method to use a single multi-layer perceptron (MLP) to represent the scene, and NICE-SLAM~\cite{niceslam} uses learnable hierarchical feature grids.  ESLAM~\cite{eslam} and Co-SLAM~\cite{coslam} further improve the scene representation with tri-planes and joint coordinate-parametric encoding. PLGSLAM~\cite{plgslam} proposes a novel progressive scene representation method which dynamically allocates new local scene representation trained with frames within a local sliding window, achieving high-accuracy scene reconstruction in large-scale scenes. MNE-SLAM~\cite{mneslam} and MCN-SLAM~\cite{mcnslam} proposes a distributed SLAM framework with implicit scene representation for accurate and efficient scene reconstruction. SNI-SLAM~\cite{snislam} further incorporates semantic features into the implicit scene representation, thereby improving accuracy. Point-SLAM~\cite{pointslam} and Loopy-SLAM~\cite{loopyslam} uses neural point clouds for the scene representation. Point-SLAM allows dynamically adapting the anchor point density to the information density of the input. NeSLAM~\cite{neslam} proposes an implicit scene representation method with depth completion and denoising network for better geometric information.  Instead
of representing maps with neural implicit features, our method utilizes the explicit 3D Gaussian representation, which can significantly improve rendering speed using splatting-based rasterization.

\noindent\textbf{GS-based SLAM.} Recently, 3D Gaussian Splatting (3DGS)~\cite{3dgs} using 3D Gaussians as primitives for real-time neural rendering. 3DGS utilizes highly optimized custom CUDA kernels
and novel algorithmic approaches, which achieve significant improvements in rendering speed without sacrificing image quality. There are many pioneer works that successfully combine the advantages of 3D Gaussian Splatting with SLAM such as SplaTAM~\cite{splatam}, GS-SLAM~\cite{gsslam}, MonoGS~\cite{monogs}, Gaussian-SLAM~\cite{gaussianslam}, RTG-SLAM~\cite{rtg-slam}, Photo-SLAM~\cite{photoslam}. Some works further improve the scene representation with semantic~\cite{sgsslam,semgauss}, loop closure~\cite{loopsplat}, and structure Manhattan world hypothesis~\cite{structureslam}.  These methods achieve fast rendering speed and high-fidelity reconstruction performance.  However, we revisit all the exisiting GS-based methods and observe that the similarity of the 3D Gaussian ellipsoids is really high. The memory and storage usage are also heavy in these GS-based SLAM systems, which makes them difficult to use in real-world scenarios and with handheld devices.

\section{Method}
The pipeline of our system is shown in Fig. \ref{fig:pipeline}. The input of our system is sequential RGB-D frames $\{I_i, D_i\}_{i=1}^M$ with known camera intrinsic $K \in R_{3\times3}$.  Our system simultaneously reconstructs a dense scene map and estimates camera poses $\{R_i|t_i\}^M_{i=1}$. For the mapping thread, a voxel-based compact 3D Gaussian scene representation (Sec.~\ref{sec:voxel}) is designed to represent the environments with sliding window-based masks (Sec.~\ref{sec:mask}) and geometry codebook (Sec.~\ref{sec:codebook}). For the camera tracking thread, a local-to-global bundle adjustment method (Sec.~\ref{sec:ba}) is designed for robust and accurate pose estimation. The network is incrementally updated with the SLAM system operation.
\subsection{Revisit GS-Based SLAM Systems}
\label{sec:scene}
Recently, 3D Gaussian Splatting (3DGS)~\cite{3dgs} has employed 3D Gaussians as the fundamental elements for real-time neural rendering. By leveraging highly optimized custom CUDA kernels, 3DGS achieves a substantial boost in rendering speed while maintaining high image quality. SplaTAM~\cite{splatam}, GS-SLAM~\cite{gsslam}, MonoGS~\cite{monogs}, Gaussian-SLAM~\cite{gaussianslam} are the pioneer works that successfully combine the advantages of 3D Gaussian Splatting with SLAM. These methods achieve fast rendering speed and high-fidelity reconstruction performance. 

However, we revisit the scene representation of these GS-based SLAM systems and observe that the geometry similarity of the 3D Gaussian ellipsoids is really high and some of them are redundant.
For all the GS-based SLAM systems, the scene representation can defined as:
\begin{equation}
    G(\mathbf{x})=o e^{-\frac{1}{2}(x-\mu)^T \boldsymbol{\Sigma}^{-1}(x-\mu)} 
    \label{eq:1}
\end{equation}
Eq.~\ref{eq:1} is the original representation of 3D Gaussian ellipsoids.
\begin{equation}  \boldsymbol{\Sigma}=\boldsymbol{R} \boldsymbol{S} \boldsymbol{S^T} \boldsymbol{R^T}
\end{equation}
The scene can be represented with a number of small Gaussian ellipsoids with 3D geometry attributes (scale and rotation matrix $\boldsymbol{S},\boldsymbol{R}$).

Then, the 3D Gaussian ellipsoids are used to render 2D images with the technique of splatting. The covariance matrix 
$\boldsymbol{\Sigma}'$ in camera coordinates can be formulated as:
\begin{equation}
\boldsymbol{\Sigma}^{\prime}=\boldsymbol{J} \boldsymbol{W} \boldsymbol{\Sigma} \boldsymbol{W}^T \boldsymbol{J}^T
\end{equation}
where $\boldsymbol{W}$ denotes the view direction, $\boldsymbol{J}$ denotes the projection transformation matrix.

\begin{figure*}[h]
  \centering
\includegraphics[width=0.95\linewidth]{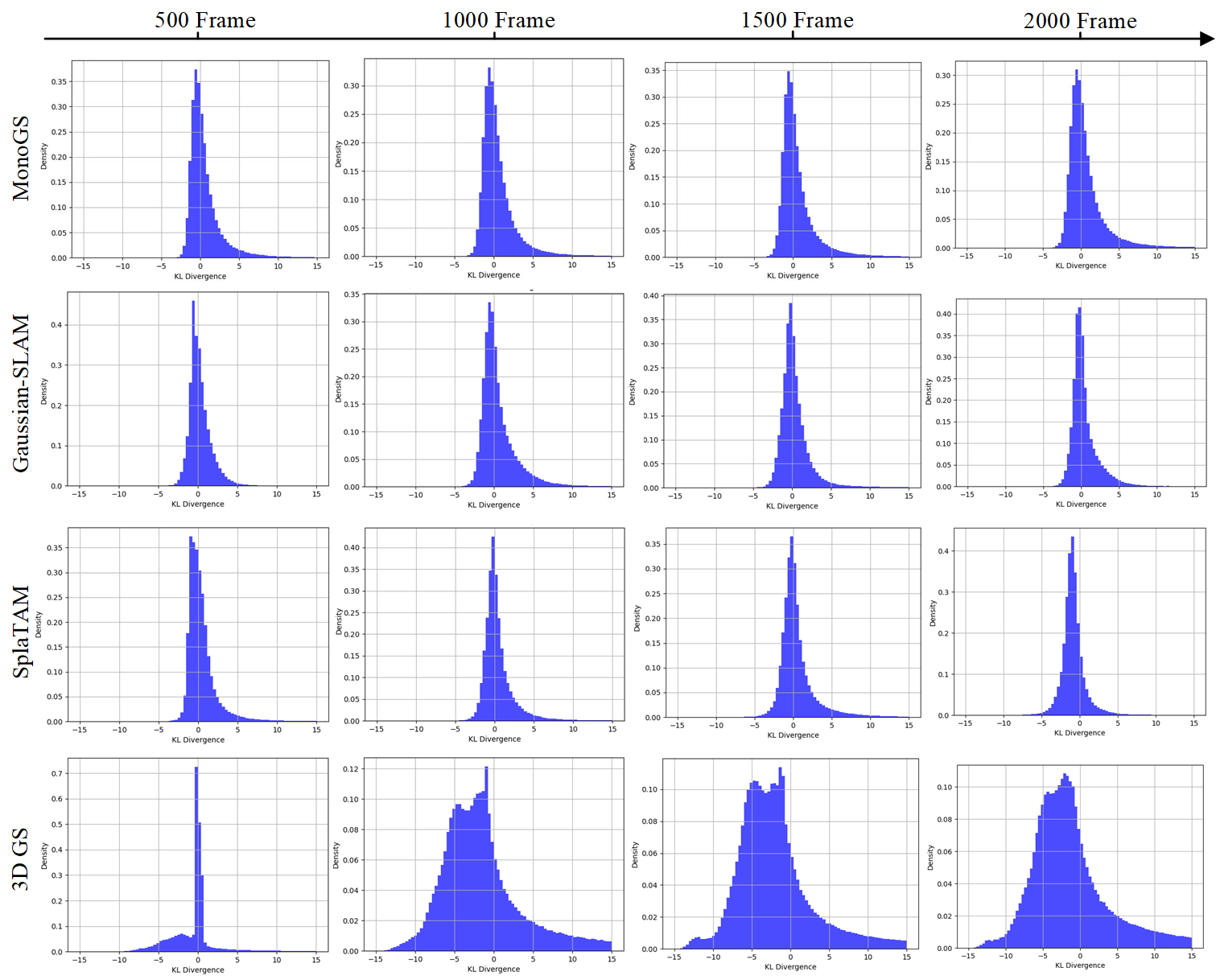}
  \caption{The KL divergence distribution of the Gaussian ellipsoids with the online training of the SLAM system on different time steps (500, 1000, 1500, 2000) in Replica dataset\cite{replica}. A larger area of the blue region signifies lower similarity between the 3D Gaussian ellipsoids, whereas a smaller area and a higher peak indicate greater similarity between the Gaussian ellipsoids. We can observe that the similarity in geometry consistently remains at a high level of GS-based SLAM system.}
  \label{fig:kl}
 
\end{figure*}

In order to analyze the geometry similarity of 3D gaussian ellipsoids $G_1$,$G_2$, we adopt the Kullback-Leibler divergence and extended it to 3D Gaussians formulation:
\begin{equation}
\begin{split}
D_{KL}(G_1(\boldsymbol{x}) \| G_2(\boldsymbol{x}))&=  \mathbb{E}_{\boldsymbol{x} \sim G_1(\boldsymbol{x})}\left[\log \frac{G_1(\boldsymbol{x})}{G_2(\boldsymbol{x})}\right] \\
&=\mathbb{E}_{\boldsymbol{x} \sim G_1(\boldsymbol{x})}\left[\log G_1(\boldsymbol{x})\right] \\ & + \mathbb{E}_{\boldsymbol{x} \sim G_1(\boldsymbol{x})}\left[-\log G_2(\boldsymbol{x})\right]
\end{split}
\end{equation}
Since both $G_1(x), G_2(x)$ follow the normal distribution, the equation can be transformed into:
\begin{equation}
\begin{split}
    D_{KL}(G_1(\boldsymbol{x}) \| G_2(\boldsymbol{x}))=&\frac{1}{2}\left[(\boldsymbol{\mu}_1-\boldsymbol{\mu}_2)^{\top} \boldsymbol{\Sigma}_2^{-1}(\boldsymbol{\mu}_1-\boldsymbol{\mu}_2) \right .\\
    & \left .-\log \operatorname{det}(\boldsymbol{\Sigma}_2^{-1} \boldsymbol{\Sigma}_1)+tr(\boldsymbol{\Sigma}_2^{-1} \right .\\
    & \left . \boldsymbol{\Sigma}_1)-n \right]
    \end{split}
\end{equation}
Then we normalized the 3D Gaussian ellipsoids $G_1$,$G_2$ to an unbiased Gaussian distribution $\mathcal{N}(0,\boldsymbol{\Sigma}_1),\mathcal{N}(0,\boldsymbol{\Sigma}_2)$, and the geometry similarities can be formulated as:
\begin{equation}
    D_{KL}=\frac{1}{2n}tr(\boldsymbol{\Sigma}_1^{-1}\boldsymbol{\Sigma}_2)-\frac{1}{2}+\frac{1}{2n}lndet(\boldsymbol{\Sigma}_1^{-1}\boldsymbol{\Sigma}_2)
\end{equation}
where n is the dimension of the covariance matrix. We conduct various benchmark and present our results on Tab.~\ref{tab:kld} and Fig.~\ref{fig:kl}. We conduct experiments on all the open-sourced GS-based SLAM systems. In Fig.~\ref{fig:kl}, we show the distribution of 3D Gaussian ellipsoids under different KL divergence values. When the KL divergence is 0, it indicates that the shapes and sizes of the two 3D Gaussian ellipsoids are identical. The larger the area covered by the curve in the figure, the more spread out the overall KL divergence value, indicating a lower similarity between the 3D Gaussian ellipsoids.  Our experiments show that the similarities of 3D Gaussians of the GS-based SLAM system are significantly greater than the original 3DGS due to the sequentialized processing. 
\begin{figure}[t]
  \centering
\includegraphics[width=\linewidth]{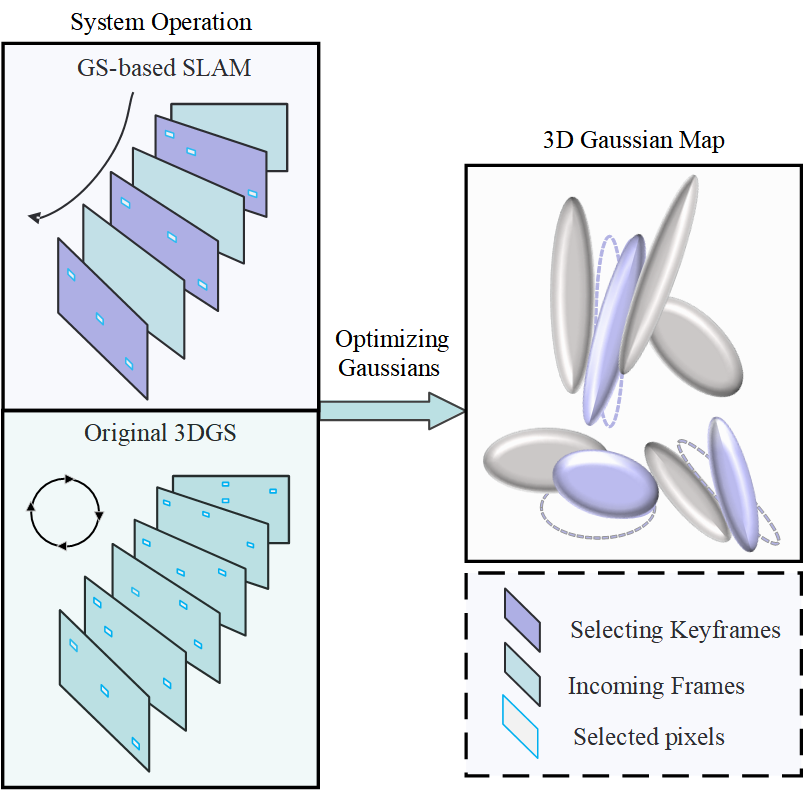}
  \caption{The core distinctions between GS-based SLAM systems and original 3DGS: the system operation (sequential/global) and keyframe selection. }
  \label{fig:optimization}
  
\end{figure}

We attempted to analyze the differences between the SLAM method and the original 3DGS method in Fig.~\ref{fig:optimization}. We present the sequentializing of dataset and online optimization process of GS-based SLAM systems and the original 3DGS. There are two core distinctions: 1) the data processing procedure. During the training process of GS-based SLAM systems, all images are input sequentially based on their timestamp, with each subsequent frame being processed only after the training of the previous frame is completed, while 3DGS directly inputs all images and performs simultaneous optimization across them. 2) the keyframe selection strategy.  For the keyframe selection strategy, the 3D Gaussian ellipsoids are constructed through an online incremental optimization scheme, where only keyframes are selectively optimized during the scene refinement process in GS-based SLAM systems. Conversely, the original GS method employs a strategy of random sampling across the entire set of images. The online optimization strategy employed by SLAM systems has, to a certain extent, exacerbated the geometric similarity of the 3D Gaussian ellipsoids within the scene.

To this end, the aforementioned experimental results conclusively demonstrate the redundancy reduction of the GS-SLAM system compared to the original 3DGS. Building upon this, we propose novel voxel-anchored 3D Gaussian representation coupled with a sliding window-based online masking mechanism specifically tailored for SLAM systems. This approach achieves substantial reductions in both the required number of Gaussian primitives,  training speed, and storage usage. Furthermore, through residual vector quantization-based model compression, we further improve the memory usage, which is crucial for real-world deployment and embedded device applications.

\begin{table*}[h]
\centering
\scalebox{0.9}{
\setlength{\tabcolsep}{1mm}{
\begin{tabular}{cc|cccc|cc}
\toprule
\multicolumn{2}{c|}{SplaTAM \cite{splatam}}  & \multicolumn{2}{c}{MonoGS \cite{monogs}}    & \multicolumn{2}{c|}{Gaussian-SLAM \cite{gaussianslam}} & \multicolumn{2}{c}{3DGS \cite{3dgs}}      \\
Range            & Percentage & Range            & Percentage & Range               & Percentage   & Range            & Percentage \\ \midrule
(-2.5\%,2.5\%)   & 87.67\%    & (-2.5\%,2.5\%)   & 83.67\%    & (-2.5\%,2.5\%)      & 88.37\%      & (-2.5\%,2.5\%)   & 20.07\%    \\
(-5.0\%,5.0\%)   & 97.86\%    & (-5.0\%,5.0\%)   & 95.86\%    & (-5.0\%,5.0\%)      & 94.45\%      & (-5.0\%,5.0\%)   & 37.04\%    \\
(-7.5\%,7.5\%)   & 99.16\%    & (-7.5\%,7.5\%)   & 98.91\%    & (-7.5\%,7.5\%)      & 97.21\%      & (-7.5\%,7.5\%)   & 47.69\%    \\
(-10.0\%,10.0\%) & 99.52\%    & (-10.0\%,10.0\%) & 99.28\%    & (-10.0\%,10.0\%)    & 99.37\%      & (-10.0\%,10.0\%) & 52.16\%    \\ \bottomrule
\end{tabular}}}
\caption{The KL divergence analysis of GS-based SLAM systems and original 3DGS.}
\label{tab:kld}
\end{table*}
\subsection{Voxel-Anchored 3D Gaussian Representation}
\label{sec:voxel}
\begin{figure*}[t]
  \centering
\includegraphics[width=\linewidth]{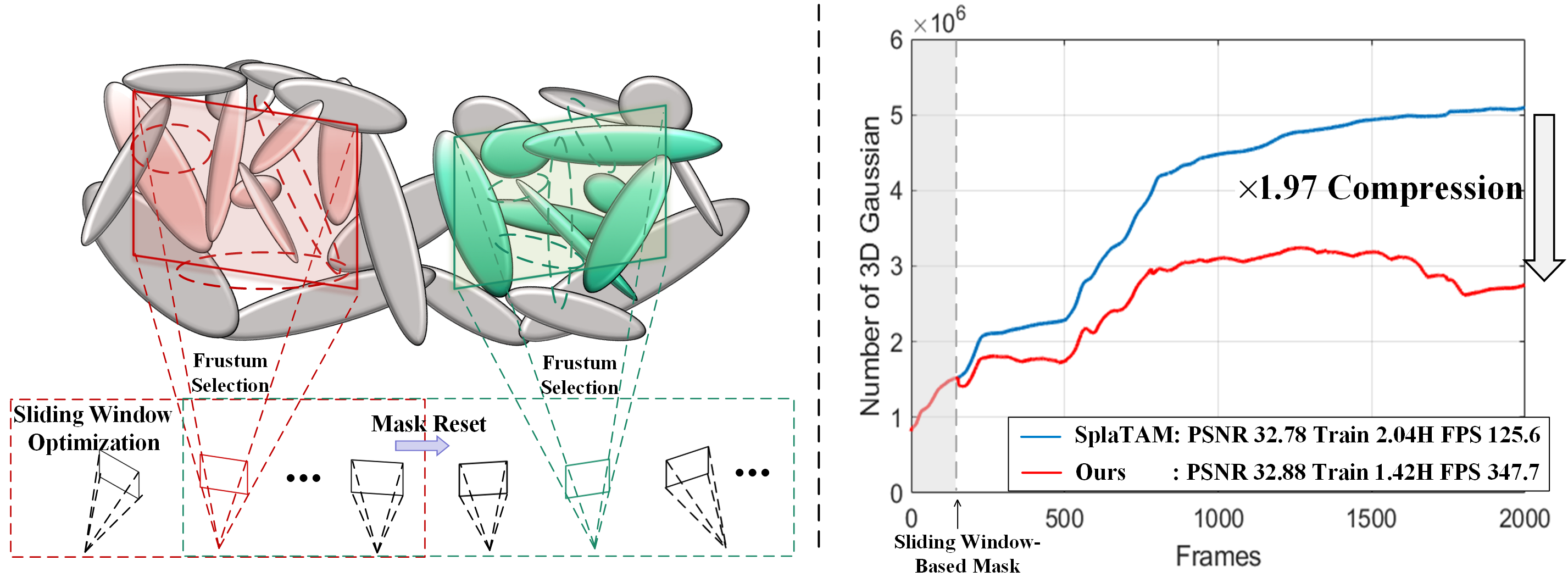}
  \caption{The left figure shows the learnable online masking strategy. We perform frustum selection and sliding widow reset to remove redundant Gaussian ellipsoids while maintaining the reconstruction accuracy efficiently. The dashed lines represent the removed 3D Gaussian ellipsoids. The right figure shows the varying count of Gaussian ellipsoids during the SLAM system operation. These two curves show the distinction between our system with and without masks. Our mask strategy achieves \textbf{ 1.97 $\times$} compression on the number of 3D Gaussians.}
  \label{fig:mask}

\end{figure*}

The existing GS-based SLAM systems directly use the original 3DGS for scene representation, achieving promising image quality. However, the original 3DGS creates a number of redundant 3D Gaussian ellipsoids with the SLAM system operation($\times$1.52 Gaussian ellipsoids show similar performance in Fig.\ref{fig:mask}), while both of these methods fail to discover this. This finally results in poor performance in training speed, memory, and storage usage, which is crucial for online SLAM systems. Some methods~\cite{compactgs1,compactgs2,lightgaussian} propose novel Gaussian pruning and self-organizing methods to compact the 3DGS attributes. However, all of these strategies are not suitable for GS-based SLAM systems as they have to obtain all the images, pose, and the corresponding point cloud at the beginning, while SLAM systems are incrementally optimized. 

Inspired by \cite{scaffoldgs}, we design a novel voxel-based sequentialized 3D Gaussian representation for compact and efficient SLAM. We utilize the scene structure prior to guide the distribution of 3D Gaussians and remove unnecessary supervisory 3D Gaussian ellipsoids, maintaining a low compute cost while avoiding unrestricted growth as the scene expands.

The scene of each frame is voxelized with the point cloud $\boldsymbol{P_i}\in \mathcal{R}^{N\times3}$. We use $\mathbf{V}\in \mathcal{R}^{N'\times3}$ to denote voxel centers. The center of each voxel is initialized as an anchor point $\textbf{x}^a\in \mathcal{R}^3$. Each anchor is characterized by its attributes $\mathcal{A}=\left\{\boldsymbol{f}^a \in \mathbb{R}^{32}, \boldsymbol{l} \in \mathcal{R}^3,\mathcal{O}\in \mathcal{R}^{k\times3}\right\}$, where each component represents the anchor feature, scaling, and offsets, respectively. Then, we derive 3D Gaussians from anchor points. The attributes of a neural Gaussian are defined as: position $\mu \in \mathcal{R}^3$, opacity $\alpha \in \mathcal{R}$, quaternion $q\in \mathcal{R}^4$, scaling $s\in \mathcal{R}^3$, and color $c \in \mathcal{R}^3$. The positions of the corresponding $k$ 3D Gaussians are calculated as:
\begin{equation}
    \{ \mu_i\}_{i=0}^{k-1} = \mathbf{x}_a+ \{ \mathcal{O}_i\}_{i=0}^{k-1} \cdot \boldsymbol{l}
\end{equation}
where $\{ \mathcal{O}_i\}_{i=0}^{k-1} \in \mathcal{R}^{k\times3}$ are the learnable offsets and $\boldsymbol{l}$ is the scaling factor associated with anchor. Then, the attributes of the Gaussians are decoded from the anchor feature $\boldsymbol{f}^a$, the viewing distance $\mathbf{\delta}$ and direction $\boldsymbol{d}$ through individual MLPs:
\begin{equation}
    \{ \boldsymbol{f}^a, \delta, \boldsymbol{d} \} \mapsto 
    \{\{ \alpha_i\}_{i=0}^{k-1}, \{ q_i\}_{i=0}^{k-1} , \{ s_i\}_{i=0}^{k-1} ,  \{ c_i\}_{i=0}^{k-1}  \}
\end{equation}
\begin{equation}
    \delta=\left\|\mathbf{x}_a-\mathbf{x}_c\right\|_2, \mathbf{d}=\frac{\mathbf{x}_a-\mathbf{x}_c}{\left\|\mathbf{x}_a-\mathbf{x}_c\right\|_2}
\end{equation}
where $\mathbf{x}_c$ denotes camera position. All attributes are decoded in a single pass, with each attribute processed by a different MLP.

\begin{figure}
    \centering
    \includegraphics[width=\linewidth]{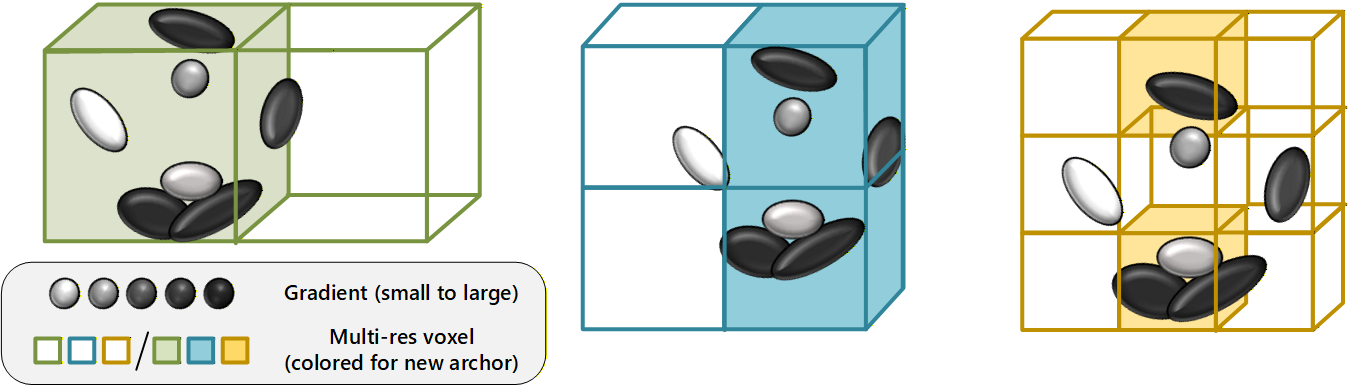}
    \caption{Growing operation. We develop an anchor growing policy guided by the gradients of the neural Gaussians. From
left to right, we spatially quantize neural Gaussians into multi-resolution voxels. New anchors
are added to voxels with aggregated gradients larger than $\tau_g$.}
    \label{fig:anchorgrow}
\end{figure}

\noindent\textbf{Sequential Anchor Point Growing.}
We have designed a sequential voxel growth strategy tailored for SLAM. In the SLAM system, keyframes are selected at fixed intervals for map construction. At the beginning of the map, we first compute a posed point cloud from the input, and then sample $M_k$ points from the regions where the accumulated $\alpha$ is below a threshold or where significant depth discrepancies occur. These points are voxelized and initialized as anchor points. Upon the arrival of each keyframe, new anchor points and 3D Gaussians are added to the map to represent newly observed regions of the scene. 
Then, for each voxel, we compute the averaged gradients of the included neural Gaussians, denoted $\nabla_g$.
During optimization, we spatially quantize the neural Gaussians by constructing voxels of a given size $\epsilon$, and compute the averaged gradients of the included Gaussians over 
N training iterations. We compute the averaged gradients of the included neural Gaussians, denoted $\nabla_g$. If the $\nabla_g > \tau_g$, this voxel is deemed as significant. $\tau_g$ is a predefined parameter. A new anchor point is deployed at the center of a voxel if none has yet been established. In practice, we quantize the space into a multi-resolution voxel grid, which allows new anchors to be added at different levels of granularity.
\begin{equation}
\epsilon^{(m)}=\epsilon_g / 4^{m-1}, \quad \tau_g^{(m)}=\tau_g * 2^{m-1}
\end{equation}
where $(m \in\{1,2,3\})$ of size $\left\{\epsilon_g^{(m)}\right\}$.
We visualize the anchor point growing strategy in Fig.~\ref{fig:anchorgrow}. The figure illustrates how new anchors are added within voxel grids of three different resolutions. If the $\nabla_g > \tau_g$, we grow new anchor points in these voxels. To further regulate the addition of new anchors, we apply a random elimination to these candidates. To remove trivial anchors, we accumulate the opacity values of their associated neural Gaussians in the map.

\subsection{Sliding Window-based Online Mask}
\label{sec:mask}

In order to further improve the compactness of our scene representation,
 we propose a learnable sliding window-based online masking strategy to remove the redundant 3D Gaussian ellipsoids tailored for the SLAM system. Compared to the original densification method, which only considers the opacity, our method takes into account both the volume $V$ and opacity $\alpha \in[0,1]$ of Gaussian ellipsoids. The volume calculation is $V=\frac{4}{3}\pi abc$, where $abc$ are the three dimensions of the scale $\boldsymbol{S}$. We introduce a learnable mask parameter $m\in R^{N_w}$ and a corresponding binary mask $M\in\{0,1\}^{N_w}$, $N_w$ is the number of Gaussian ellipsoids within the sliding window:

 \begin{equation}
M_n=\operatorname{sg}\left(\mathbb{I}\left[Sig\left(m_n\right)>\epsilon\right]-Sig\left(m_n\right)\right)+Sig\left(m_n\right)
 \end{equation}

\begin{equation}
    \hat{s_n}=M_n s_n, \qquad \hat{\alpha_n}=M_n\alpha_n
\end{equation}
where $n$ is the index of the Gaussian ellipsoids, $\epsilon$ denotes the mask threshold. Inspired by~\cite{gradient,compactgs1}, we employ the stop gradient operator $sg(\cdot)$ to calculate gradients from binary masks. $\mathbb{I}$ and $Sig(\cdot)$ denote the indicator and sigmoid function. This formulation of mask strategy allows us to effectively combine the influence of volume ($\hat{s_n}$) and opacity($\hat{o_n}$) of Gaussian ellipsoids. We utilize $\hat{s_n}$, $\hat{\alpha_n}$ in the color and depth rendering. Then, We formulate the loss function $L_m$ of our mask:
\begin{equation}
    L_m=\frac{1}{N_w}\sum_{n=1}^{N_w} Sig(m_n), \quad h=I(G(\boldsymbol{x}),r_i)
\end{equation}

 $h$ represents the hit count of each 3D Gaussian within a sliding window. In order to better fit the online updating SLAM systems, we further improve the masking strategy by adding frustum culling and sliding window-based reset strategy, shown in Fig.~\ref{fig:mask}. Our frustum culling strategy allows us to optimize only the mask within the current viewing frustum while keeping the rest of the 3D Gaussian ellipsoids fixed. It will not only preserve the previously reconstructed geometry but also significantly reduce the number of parameters during optimization. Different from the original densification strategy performed on every frame, we only perform mask on the keyframe (each $k^{th}$ frame) for efficiency and accuracy.

 Compared to the original 3DGS method, for the SLAM system, we need to consider the sequential characteristic of the system when creating masks. Therefore, we design a local sliding window online mask method. The optimization process of all masks is synchronized with the system's operation. We maintain a local sliding window mask and perform a sliding window reset to avoid the continuous optimization and accumulated gradient of masks which will ultimately eliminate all Gaussian ellipsoids. The sliding window consists of the current frame, the most relevant keyframe, and $n-2$ previous keyframes, which have the highest overlap with the current frame. Overlap is evaluated by analyzing the point cloud of the current frame's depth map and tallying points within the frustum of each keyframe. This can also ensure the consistency of the mask within the local sliding window.  We calculate the hit count of each
Gaussian. When moving to the bound of the sliding window, we remove the redundant Gaussian ellipsoids by evaluating their hit frequency in conjunction with the learned mask.
 This approach allows us to continuously mask out unnecessary Gaussians during online SLAM system operation, effectively reducing computation overhead and ensuring efficient memory usage on GPU.

\subsection{Residual Codebook Quantization for Neural Anchor Point}
\label{sec:codebook}

In this section, we propose a learnable codebook and employ a residual vector quantization method to reduce computational complexity and memory usage and further improve the training and rendering speed.
\begin{figure*}[t]
  \centering
\includegraphics[width=\linewidth]{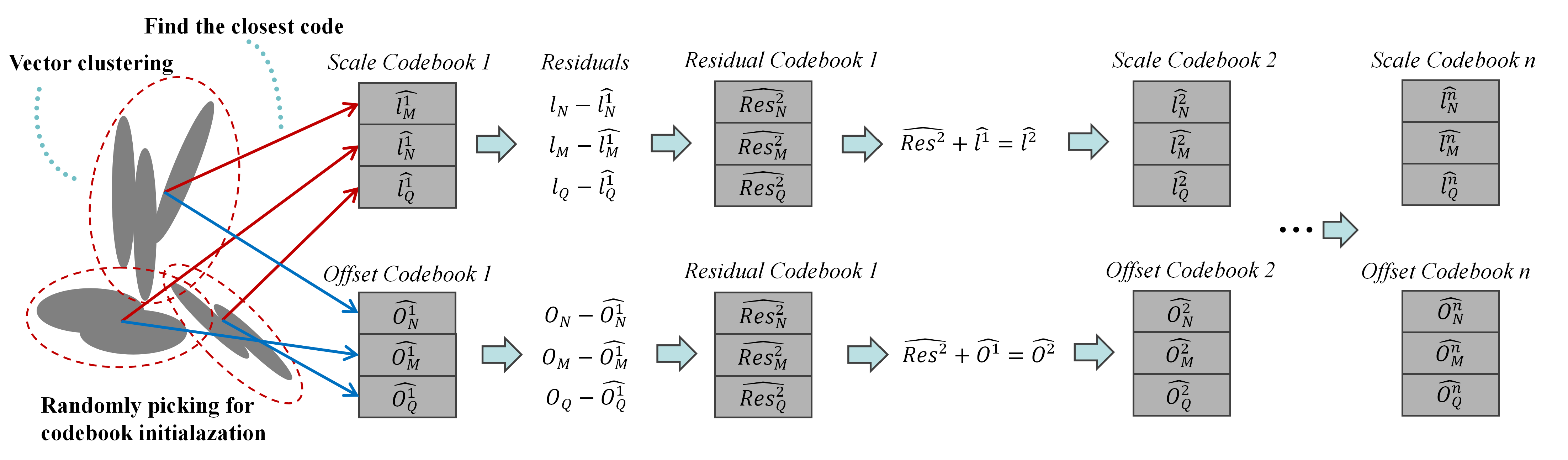}
  \caption{The R-VQ process to represent the attributes of neural anchor points. In the first stage, we cluster the offset and scaling factor vector and randomly select codebook initialization with the closest code. In the subsequent stage, the residual between the original vector and the result from the first stage is stored in another codebook. This iterative process continues through to the ultimate stage, at which point, the collectively chosen indices and codebook from each stage provide a representation of the original vector.}
  \label{fig:rvq}

\end{figure*}

Based on the similarity, we propose a learnable codebook to compress the geometry and feature attributes, shown in Fig. \ref{fig:rvq}. Inspired by \cite{soundstream,compactgs1,lightgaussian}, we incorporate residual vector quantization (R-VQ) to compress the offset $\mathcal{O}$ and scaling factor $\boldsymbol{l}$. It cascades L stages of VQ and is formulated as follows:
\begin{equation}
\begin{aligned}
\hat{\boldsymbol{l}}_n^l & =\sum_{k=1}^l \mathcal{C}^k\left[i^k\right], \quad l \in\{1, \ldots, L\}, \\
i_n^l & =\underset{k}{\operatorname{argmin}}\left\|\mathcal{C}^l[k]-\left(\boldsymbol{l}_n-\hat{\boldsymbol{l}}_n^{l-1}\right)\right\|_2^2, \quad \hat{\boldsymbol{l}}_n^0=\overrightarrow{0}
\end{aligned}
\end{equation}
where $\boldsymbol{l}\in\mathcal{R}^{N\times3}$ is the scale vector, $\hat{\boldsymbol{l}}^l\in\mathcal{R}^{N\times 3}$ is the output scale vector after $l$ stages quantization. $n$ denotes the index of the Gaussian ellipsoids. $\mathcal{C}^l$ denotes the codebook at the stage l. $\mathcal{C}^l$ represents the vector at index i of the codebook $\mathcal{C}$. The formulation of the rotation vector is the same. Then, the loss function is defined as:
\begin{equation}
    L_{\boldsymbol{l}}=\frac{1}{N P} \sum_{k=1}^L \sum_{n=1}^N\left\|\operatorname{sg}\left[\boldsymbol{l}_n-\hat{\boldsymbol{l}}_n^{k-1}\right]-\mathcal{C}^k\left[i_n^k\right]\right\|_2^2
\end{equation}
where $P$ is the size of codebook, $sg[\cdot]$ is the stop gradient operator. The codebook size is set to 64, and the stage is set to 6. After this, we can only store the codebook compressed offset and scaling factor, which can significantly reduce storage and memory usage.

\begin{figure*}[t]
  \centering
\includegraphics[width=\linewidth]{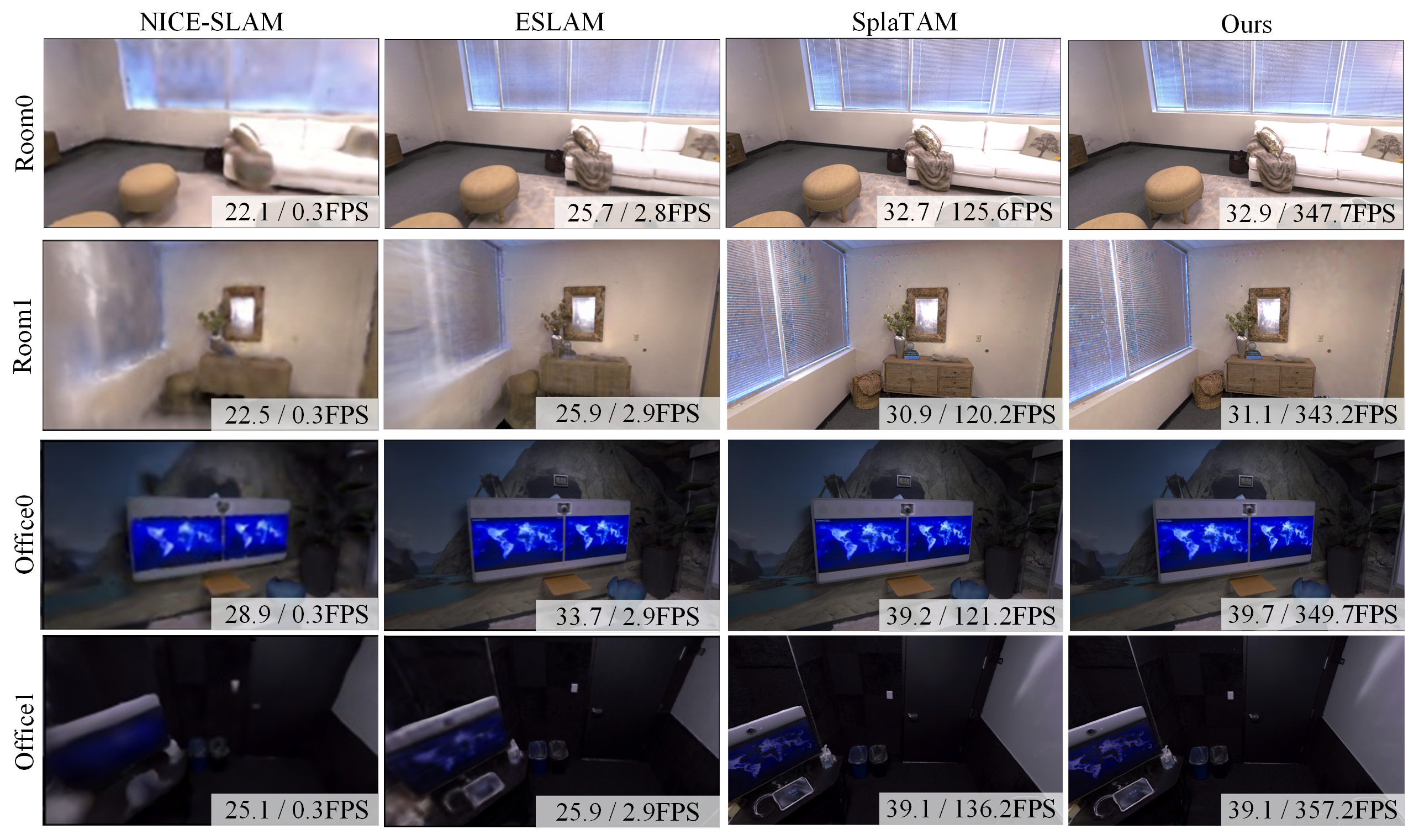}
  \caption{The rendering visualization results on the Replica dataset~\cite{replica} of the proposed GS-based SLAM system compared with other SOTA methods. We present the rendering PSNR and FPS on the image. Our method can
achieve faster rendering speed and high-quality image reconstruction performance compared with other methods.}
  \label{fig:replica}

\end{figure*}

\begin{table*}[]
\centering
\scalebox{0.8}{
\setlength{\tabcolsep}{1.0mm}{
\begin{tabular}{llllll}
\toprule
Modality                 & Hardware & ROS Topic  & Type                                                                        & Description                                                                   & Rate                   \\ \midrule
                         &                              & /camera/depth/image\_rect\_raw &                                                                             & 848×480 Depth                                                              &                        \\
                         &      & /camera/infra1/image\_rect\_raw  &                                                                             & 640×480 Greyscale                                                             &  \\
                         & \multirow{-3}{*}{D435i}      & /camera/infra2/image\_rect\_raw &                                                                             & 640x480 Greyscale                                                             &                        \\
                         &                              & /camera/color/image\_raw        & \multirow{-4}{*}{sensor\_msgs/Image}                                        & 640×480 RGB                                                                   & \multirow{-4}{*}{30HZ}   \\   
\multirow{-4}{*}{Camera}    
                         & D435i                        & /d435i/imu                     &                                                                             & Bosch BMI055                                                                  & 400HZ                  \\
                                        
\multirow{-2}{*}{IMU}    & Livox Mid360                 & /livox/lidar/imu               & \multirow{-3}{*}{sensor\_msgs/Imu}                                          &                                                                               & 200HZ                  \\
Lidar                    & Livox Mid360                 & /livox/lidar                   & livox\_ros\_driver/CustomMsg & \begin{tabular}[c]{@{}l@{}}1 channel.\\ Points per channel: 9984\end{tabular} & 10HZ                   \\                    & HAP Lidar                & /livox/lidar\_192\_168\_1\_3                  & livox\_ros\_driver/CustomMsg &  & 10HZ                   \\ \hline
\end{tabular}}}
\caption{ Summary of sensing modalities, hardware units, ROS topics, and the nominal rates on each platform in our dataset. All these data have also been directly extracted from the rosbag and saved as individual files.}
\label{tab:sensor}
\end{table*}

\begin{table*}[t]
\centering

\scalebox{0.90}{
\setlength{\tabcolsep}{1.5mm}{
\begin{tabular}{lccccccccc}
\toprule
Methods    & Average & Room0   & Room1   & Room2   & Office0  & Office1  & Office2  & Office3  & Office4  \\ \midrule
\multicolumn{10}{l}{\cellcolor[HTML]{EEEEEE}{\textit{NeRF-Based Methods}}} \\ \cdashline{1-10}
Vox-Fusion~\cite{voxfusion} & 3.09 & 1.37 & 4.70 & 1.47 & 8.48 & 2.04 & 2.58 & 1.11 & 2.94 \\
NICE-SLAM~\cite{niceslam}  & 1.06 & 0.97 & 1.31 & 1.07 & 0.88 & 1.00 & 1.06 & 1.10 & 1.13 \\
ESLAM~\cite{eslam}      & 0.63 & 0.71 & 0.70 & 0.52 & 0.57 & 0.55 & 0.58 & 0.72 & 0.63 \\
Co-SLAM~\cite{coslam}     & 0.78 & 0.61 & 0.79 & 0.96 & 0.54 & 0.52 & 1.13 & 0.97 & 0.74 \\
Point-SLAM~\cite{pointslam} & 0.52 & 0.61 & 0.41 & 0.37 & 0.38 & 0.48 & 0.54 & 0.69 & 0.72 \\
PLGSLAM~\cite{plgslam}               & 0.57          & 0.64                     & 0.65                     & 0.49                     & 0.51  & 0.52  & 0.54  & 0.65  & 0.57   \\
Loopy-SLAM~\cite{loopyslam}   & 0.30   & \cellcolor{tabthird}0.29 & 0.24  & 0.28  & 0.28  & 0.42 & 0.30  & 0.23  & 0.35   \\ 
\multicolumn{10}{l}{\cellcolor[HTML]{EEEEEE}{\textit{3DGS-Based Methods}}} \\ \cdashline{1-10}
SplaTAM~\cite{splatam}    & 0.34 & 0.31 & 0.40 & 0.29 & 0.47 & 0.27 & 0.29 & 0.32 & 0.55 \\ MonoGS~\cite{monogs}    & 0.37 & 0.33 & 0.35 & 0.31 & 0.45 & 0.29 & \cellcolor{tabthird}0.28 & \cellcolor{tabthird}0.28 & 0.75 \\
Gaussian-SLAM~\cite{gaussianslam}    & 0.35 & \cellcolor{tabthird}0.29 & 0.34 & \cellcolor{tabsecond}0.25 & 0.43 & 0.26 & 0.41 & 0.33 & \cellcolor{tabthird}0.47 \\
Photo-SLAM\cite{photoslam}      & 0.57                      & 0.64                     & 0.65                     & 0.49                     & 0.51  & 0.52  & 0.54  & 0.65  & 0.57   \\
\hdashline
\cellcolor{gray!20}Ours(Jetson)       & \cellcolor{tabthird}0.32 & \cellcolor{tabthird}0.29 & \cellcolor{tabthird}0.31 & 0.26 & \cellcolor{tabthird}0.42 & \cellcolor{tabthird}0.24 & 0.29 & 0.31 & 0.48 \\ 
\cellcolor{gray!20}Ours(Laptop)       & \cellcolor{tabsecond}0.29 & \cellcolor{tabsecond}0.26 & \cellcolor{tabsecond}0.28 & \cellcolor{tabsecond}0.25 & \cellcolor{tabsecond}0.39 & \cellcolor{tabsecond}0.21 & \cellcolor{tabsecond}0.27 & \cellcolor{tabsecond}0.26 & \cellcolor{tabsecond}0.41 \\
\cellcolor{gray!20}Ours       & \cellcolor{tabfirst}0.27 & \cellcolor{tabfirst}0.24 & \cellcolor{tabfirst}0.24 & \cellcolor{tabfirst}0.23 & \cellcolor{tabfirst}0.32 & \cellcolor{tabfirst}0.20 & \cellcolor{tabfirst}0.25 & \cellcolor{tabfirst}0.25 & \cellcolor{tabfirst}0.39 \\
\bottomrule
\end{tabular}}}
\caption{Camera tracking results on Replica dataset~\cite{replica}. We use the ATE RMSE $\downarrow [cm]$ as the metric and compare it with other SOTA methods. Best results are highlighted as \colorbox{tabfirst}{first}, \colorbox{tabsecond}{second}, and \colorbox{tabthird}{third}.}
\label{tab:replica}
\end{table*}
\begin{table}[t]
\centering

\scalebox{0.84}{
\setlength{\tabcolsep}{0.95mm}{
\begin{tabular}{lccccccc}
\toprule
Methods    & Avg.  & 0000  & 0059  & 0106  & 0169  & 0181  & 0207 \\ \midrule
\multicolumn{8}{l}{\cellcolor[HTML]{EEEEEE}{\textit{NeRF-Based Methods}}} \\ \cdashline{1-8}
Vox-Fusion~\cite{voxfusion} & 26.90 & 68.84 & 24.18 & 8.41  & 27.28 & 23.30 & 9.41 \\
NICE-SLAM~\cite{niceslam}  & 10.88 & 12.00 & 14.00 & 7.90  & 10.90 & 13.40 & \cellcolor{tabsecond}6.20 \\
Co-SLAM~\cite{coslam}    & \cellcolor{tabsecond}8.72  & \cellcolor{tabfirst}7.35  & 11.45 & \cellcolor{tabthird}9.68  & \cellcolor{tabfirst}6.03  & 11.93 & 8.86 \\
ESLAM~\cite{eslam}      & \cellcolor{tabfirst}7.82  & \cellcolor{tabsecond}7.84  & 9.24  & \cellcolor{tabfirst}7.82  & \cellcolor{tabsecond}6.78  & \cellcolor{tabfirst}9.35  & \cellcolor{tabfirst}5.88 \\
Point-SLAM~\cite{pointslam} & 12.19 & \cellcolor{tabthird}10.24 & \cellcolor{tabfirst}7.81  & \cellcolor{tabsecond}8.65  & 22.16 & 14.77 & 9.54 \\
\multicolumn{8}{l}{\cellcolor[HTML]{EEEEEE}{\textit{3DGS-Based Methods}}} \\ \cdashline{1-8}
SplaTAM~\cite{splatam}    & 11.88 & 12.83 & 10.10 & 17.72 & 12.08 & 11.10 & 7.46 \\
MonoGS~\cite{monogs}    & 10.81 & 10.89 & 19.10 & 11.72 & 10.01 & 19.10 & 7.96 \\
Gaussian-SLAM~\cite{gaussianslam}    & 11.97 & 13.53 & 11.36 & 16.47 & 11.78 & 11.43 & 7.34 \\
\cellcolor{gray!20}Ours(Jetson)       & 10.84 & 11.74 & 9.36  & 16.87 & 11.43 & 10.98 & 6.78\\
\cellcolor{gray!20}Ours(Laptop)       & 10.79 & 11.31 & \cellcolor{tabthird}9.05  & 15.98 & 11.01 & \cellcolor{tabsecond}10.85 & 6.54 \\
\cellcolor{gray!20}Ours      & \cellcolor{tabthird}10.57 & 10.81 & \cellcolor{tabsecond}9.01  & 15.57 & \cellcolor{tabthird}10.83 & \cellcolor{tabthird}10.73 & \cellcolor{tabthird}6.41 \\
\bottomrule
\end{tabular}}}
\caption{Camera tracking results on ScanNet datasets~\cite{scannet1}. We use the ATE RMSE $\downarrow [cm]$ as the metric and compare it with other SOTA methods.}
\label{tab:scannet}
\end{table}

\begin{figure*}[t]
  \centering
\includegraphics[width=0.95\linewidth]{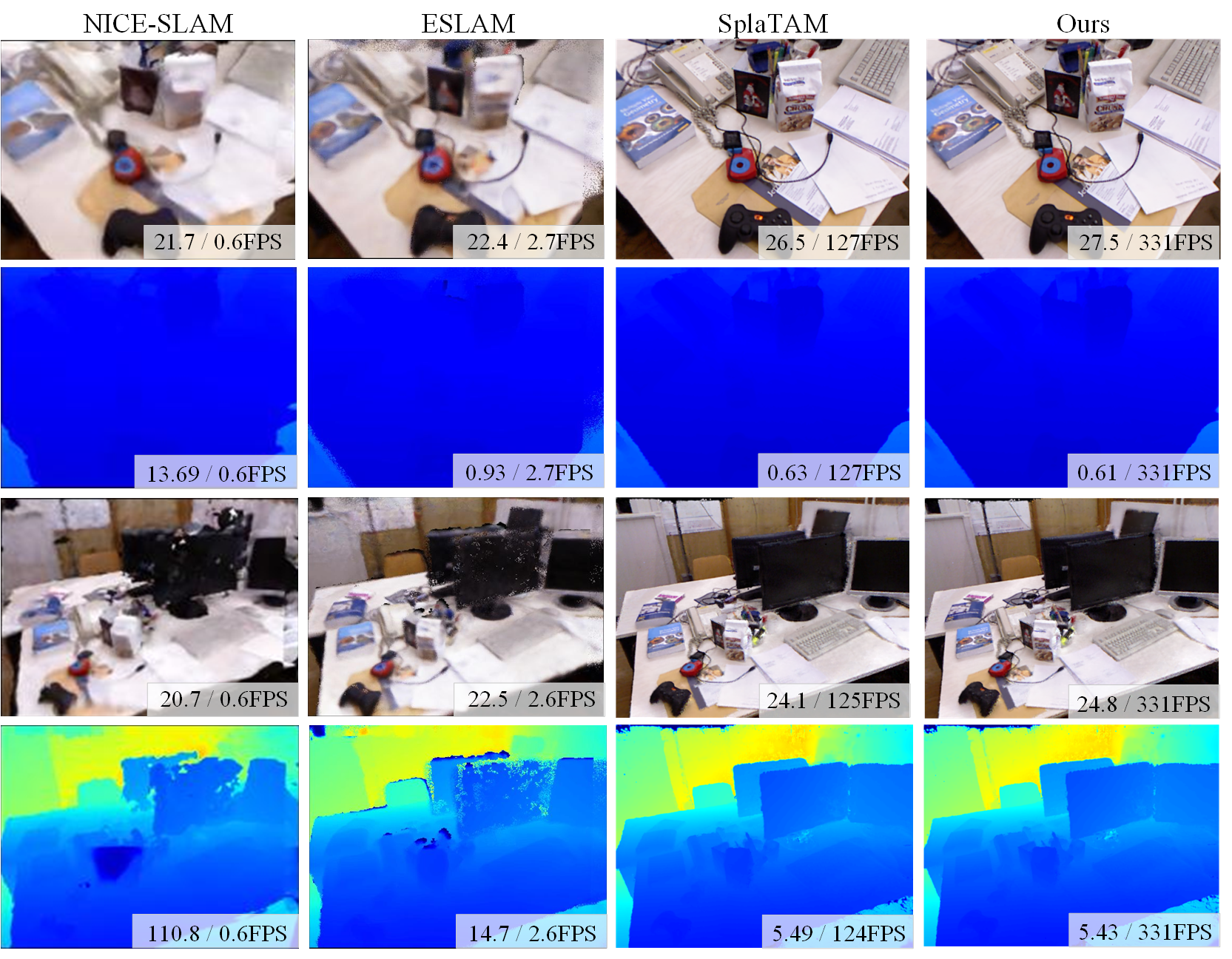}
  \caption{The rendering visualization results on the TUM RGB-D dataset~\cite{tum} of the proposed GS-based SLAM system compared with other SOTA methods. We present the rendering PSNR/FPS on the RGB image and depth L1/FPS [cm] on the depth image. Our method can
achieve faster rendering speed and high-quality image reconstruction performance compared with other methods.}
  \label{fig:tum}
\end{figure*}

\begin{table}[t]
\centering
\scalebox{0.80}{
\setlength{\tabcolsep}{0.8mm}{
\begin{tabular}{lcccccc}
\toprule
Methods       & Avg.  & fr1/desk & fr1/desk2 & fr1/room & fr2/xyz & fr3/office \\ \midrule
\multicolumn{7}{l}{\cellcolor[HTML]{EEEEEE}{\textit{Traditional Methods}}} \\ \cdashline{1-7}
Kintinuous~\cite{Kintinuous}    & 4.84  & 3.70     & 7.10      & 7.50     & 2.90    & 3.00     \\
ElasticFusion~\cite{elasticfusion} & 6.91  & 2.53     & 6.83      & 21.49    & 1.17    & 2.52     \\
ORB-SLAM2~\cite{orbslam2}     & \cellcolor{tabfirst}1.98  & \cellcolor{tabfirst}1.60     & \cellcolor{tabfirst}2.20      & \cellcolor{tabfirst}4.70     & \cellcolor{tabfirst}0.40    & \cellcolor{tabfirst}1.00     \\ 
\multicolumn{7}{l}{\cellcolor[HTML]{EEEEEE}{\textit{NeRF-Based Methods}}} \\ \cdashline{1-7}
NICE-SLAM~\cite{niceslam}     & 15.87 & 4.26     & 4.99      & 34.49    & 31.73   & 3.87     \\
ESLAM~\cite{eslam} &7.95   &2.47 &3.69  &  29.79  &  1.41   & 2.41 \\
Vox-Fusion~\cite{voxfusion}    & 11.31 & 3.52     & 6.00      & 19.53    & 1.49    & 26.01    \\
Point-SLAM~\cite{pointslam}    & 8.92  & 4.34     & 4.54      & 30.92    & \cellcolor{tabsecond}1.31    & 3.48     \\ \multicolumn{7}{l}{\cellcolor[HTML]{EEEEEE}{\textit{NeRF-Based Methods}}} \\ \cdashline{1-7}
SplaTAM~\cite{splatam}       & 5.48  & 3.35     & 6.54      & 11.13    & 1.24    & 5.16     \\ 
MonoGS~\cite{monogs}       & 3.96 & 1.59     & 6.68      & 8.55    & 1.44    & 1.52     \\ 
Gaussian-SLAM~\cite{gaussianslam}       & 5.36   & 2.74     & \cellcolor{tabsecond}6.05     & 11.32    & 1.39    & 5.33     \\ 
\cellcolor{gray!20}Ours(Jetson)          & \cellcolor{tabthird}3.98  & 1.64    & 6.65       & 8.47    & 1.41    & 1.54    \\
\cellcolor{gray!20}Ours(Laptop)          & 3.81  & \cellcolor{tabthird}1.58     & 6.27      & \cellcolor{tabthird}8.31    & 1.39    & \cellcolor{tabsecond}1.49    \\
\cellcolor{gray!20}Ours          & \cellcolor{tabsecond}3.79  & \cellcolor{tabsecond}1.57     & \cellcolor{tabthird}6.24      & \cellcolor{tabsecond}8.29    & \cellcolor{tabthird}1.37    & \cellcolor{tabsecond}1.49     \\
\bottomrule
\end{tabular}}}
\caption{Camera tracking results on TUM RGB-D dataset~\cite{tum}. We use the ATE RMSE $\downarrow [cm]$ as the metric and compare it with other SOTA methods.}
\label{tab:tum}
\end{table}

\begin{table*}[t]
\centering

\scalebox{0.80}{
\begin{tabular}{llccccccccc}
\hline
Methods                       & Metrics & \multicolumn{1}{l}{Average} & Room0    & Room1    & Room2    & Office0   & Office1   & Office2   & Office3   & Office4   \\ \hline \multicolumn{11}{l}{\cellcolor[HTML]{EEEEEE}{\textit{NeRF-Based Methods}}} \\ \cdashline{1-11}
\multirow{3}{*}{Co-SLAM \cite{coslam}}      & PSNR$\uparrow$    & 30.24                    & 27.27 & 28.45 & 29.06 & 34.14 & 34.87 & 28.43 & 28.76 & 30.91 \\
                              & SSIM$\uparrow$    & 0.94                     & 0.91  & 0.91  & 0.93  & 0.96  & 0.97  & 0.94  & 0.94  & 0.96  \\
                              & LPIPS$\downarrow$  & 0.25                     & 0.32  & 0.29  & 0.27  & 0.21  & 0.20  & 0.26  & 0.23  & 0.24  \\ \hline
\multirow{3}{*}{ESLAM \cite{eslam}}        & PSNR$\uparrow$    & 29.08                    & 25.32 & 27.77 & 29.08 & 33.71 & 30.20 & 28.09 & 28.77 & 29.71 \\
                              & SSIM$\uparrow$    & 0.93                     & 0.86  & 0.90  & 0.93  & 0.96  & 0.92  & 0.94  & 0.95  & 0.95  \\
                              & LPIPS$\downarrow$ & 0.25                     & 0.31  & 0.30  & 0.25  & 0.18  & 0.23  & 0.24  & 0.20  & 0.20  \\ \hline
\multirow{3}{*}{NICE-SLAM \cite{niceslam}}    & PSNR$\uparrow$    & 24.42                    & 22.12 & 22.47 & 24.52 & 29.07 & 30.34 & 19.66 & 22.23 & 24.49 \\
                              & SSIM$\uparrow$    & 0.81                     & 0.69  & 0.76  & 0.81  & 0.87  & 0.89  & 0.80  & 0.80  & 0.86  \\
                              & LPIPS$\downarrow$   & 0.23                     & 0.33  & 0.27  & 0.21  & 0.23  & 0.18  & 0.24  & 0.21  & 0.20  \\ \hline
\multirow{3}{*}{Point-SLAM \cite{pointslam}}   & PSNR$\uparrow$    & 35.17                    & 32.40 & 34.08 & 35.50 & 38.26 & 39.16 & 33.99 & 33.48 & 33.49 \\
                              & SSIM$\uparrow$    & \cellcolor{tabfirst}0.98                     & \cellcolor{tabthird}0.97  & \cellcolor{tabfirst}0.98  & \cellcolor{tabfirst}0.98  & \cellcolor{tabfirst}0.98  & \cellcolor{tabfirst}0.99  & \cellcolor{tabthird}0.96  & \cellcolor{tabfirst}0.96  & \cellcolor{tabfirst}0.98  \\
                              & LPIPS$\downarrow$   & 0.12                     & 0.11  & 0.12  & 0.11  & 0.10  & 0.12  & 0.16  & 0.13  & 0.14  \\ \hline 
                              \multirow{3}{*}{GO-SLAM~\cite{goslam}}                
                                                                       & PSNR$\uparrow$ & 24.42  & 24.32 & 25.37 & 26.52 & 30.17 & 30.34 & 24.16 & 28.23 & 27.64  \\
                                                                       & SSIM$\uparrow$ & 0.898  & 0.854 & 0.871 & 0.905 & 0.904 & 0.893 & 0.912 & 0.927 & 0.918  \\
                                                                       & LPIPS$\downarrow$   & 0.21                     & 0.30  & 0.27  & 0.23  & 0.21  & 0.17  & 0.23  & 0.20  & 0.18 \\
                              \multicolumn{11}{l}{\cellcolor[HTML]{EEEEEE}{\textit{3DGS-Based Methods}}} \\ \cdashline{1-11}
\multirow{3}{*}{SplaTAM \cite{splatam}}      & PSNR$\uparrow$    & 34.11                    & 32.86 & 33.89 & 35.25 & 38.26 & 39.17 & 31.97 & 29.70 & 31.81 \\
                              & SSIM$\uparrow$    & 0.97                     & \cellcolor{tabsecond}0.98  & 0.97  & \cellcolor{tabfirst}0.98  & \cellcolor{tabfirst}0.98  & 0.97  & \cellcolor{tabsecond}0.96  & 0.95  & 0.95  \\
                              & LPIPS$\downarrow$   & 0.10                     & \cellcolor{tabfirst}0.07  & 0.10  & \cellcolor{tabfirst}0.08  & 0.09  & 0.09  & 0.10  & 0.12  & 0.15  \\ \hline
\cellcolor{gray!20} &\cellcolor{gray!20}PSNR$\uparrow$     & 34.31                    & 32.99 & 34.13 & 35.42 & 38.36 & 39.47 & 32.58 & 31.21 & 32.46 \\
                              \cellcolor{gray!20}& \cellcolor{gray!20}SSIM$\uparrow$    & \cellcolor{tabfirst}0.98                     & \cellcolor{tabfirst}0.99  & \cellcolor{tabfirst}0.98  & \cellcolor{tabfirst}0.98  & \cellcolor{tabfirst}0.98  & \cellcolor{tabsecond}0.98  & \cellcolor{tabfirst}0.98  & \cellcolor{tabfirst}0.96  & \cellcolor{tabsecond}0.96  \\ \multirow{-3}{*}{\cellcolor{gray!20}SplaTam+Ours}
                              & \cellcolor{gray!20}LPIPS$\downarrow$  & 0.09                     & \cellcolor{tabfirst}0.07  & 0.10  & \cellcolor{tabfirst}0.08  & 0.09  & 0.09  & 0.10  & 0.12  & 0.15  \\ \hline
\multirow{3}{*}{MonoGS \cite{monogs}}       & PSNR$\uparrow$    & 37.24                    & \cellcolor{tabthird}34.51 & \cellcolor{tabthird}36.27 & 37.27 & \cellcolor{tabsecond}39.45 & \cellcolor{tabthird}41.98 & 36.07 & 36.45 & 35.88 \\
                              & SSIM$\uparrow$    & \cellcolor{tabthird}0.97                     & \cellcolor{tabthird}0.97  & 0.97  & 0.97  & 0.97  & \cellcolor{tabsecond}0.98  & \cellcolor{tabthird}0.96  & \cellcolor{tabfirst}0.96  & 0.95  \\
                              & LPIPS$\downarrow$   & \cellcolor{tabthird}0.080                    & 0.08  & \cellcolor{tabsecond}0.08  & \cellcolor{tabfirst}0.08  & \cellcolor{tabfirst}0.07  & \cellcolor{tabfirst}0.07  & \cellcolor{tabfirst}0.08  & \cellcolor{tabsecond}0.07  & \cellcolor{tabsecond}0.11  \\ \hline
\cellcolor[gray]{0.9} & \cellcolor{gray!20}PSNR$\uparrow$    & \cellcolor{tabthird}37.45                    & \cellcolor{tabsecond}34.69 & \cellcolor{tabthird}36.34 & \cellcolor{tabsecond}37.31 & \cellcolor{tabthird}39.27 & \cellcolor{tabsecond}42.05 & \cellcolor{tabthird}36.28 & \cellcolor{tabthird}36.72 & \cellcolor{tabthird}36.07 \\
                              \cellcolor{gray!20} & \cellcolor{gray!20}SSIM$\uparrow$    & \cellcolor{tabfirst}0.98                     & \cellcolor{tabthird}0.97  & \cellcolor{tabfirst}0.98  & \cellcolor{tabfirst}0.98  & 0.97  & \cellcolor{tabsecond}0.98  & \cellcolor{tabthird}0.96  & \cellcolor{tabfirst}0.96  & \cellcolor{tabsecond}0.96  \\
                              \multirow{-3}{*}{MonoGS+Ours}
                        \cellcolor{gray!20} & \cellcolor{gray!20}LPIPS$\downarrow$  & \cellcolor{tabfirst}0.075                    & \cellcolor{tabfirst}0.07  & \cellcolor{tabsecond}0.08  & \cellcolor{tabfirst}0.08  & \cellcolor{tabfirst}0.07  & \cellcolor{tabfirst}0.07  & \cellcolor{tabfirst}0.08  & \cellcolor{tabfirst}0.06  & \cellcolor{tabfirst}0.10  \\ \hline 
                              
                              \multirow{3}{*}{Gaussian-SLAM~ \cite{gaussianslam}}       & PSNR$\uparrow$    & \cellcolor{tabsecond}37.51                    & \cellcolor{tabfirst}34.78 & \cellcolor{tabsecond}36.83 & \cellcolor{tabsecond}37.31 & 39.15 & 41.68 & \cellcolor{tabsecond}36.85 & \cellcolor{tabsecond}37.15 & \cellcolor{tabsecond}36.38 \\
                              &SSIM$\uparrow$    & \cellcolor{tabthird}0.97                     & \cellcolor{tabthird}0.97  & 0.97  & 0.97  & \cellcolor{tabfirst}0.98  & \cellcolor{tabsecond}0.98  & \cellcolor{tabthird}0.96  & \cellcolor{tabfirst}0.96  & \cellcolor{tabsecond}0.96  \\
                              & LPIPS$\downarrow$   & \cellcolor{tabsecond}0.077                    & 0.08  & \cellcolor{tabfirst}0.07  & \cellcolor{tabfirst}0.08  & \cellcolor{tabfirst}0.07  & \cellcolor{tabfirst}0.07  & \cellcolor{tabfirst}0.08  & \cellcolor{tabsecond}0.07  & \cellcolor{tabsecond}0.11  \\ \hline
\cellcolor[gray]{0.9} & \cellcolor{gray!20}PSNR$\uparrow$    & \cellcolor{tabfirst}37.81                    & \cellcolor{tabsecond}34.69 & \cellcolor{tabfirst}37.31 & \cellcolor{tabfirst}37.64 & \cellcolor{tabfirst}39.75 & \cellcolor{tabfirst}42.49 & \cellcolor{tabfirst}37.06 & \cellcolor{tabfirst}37.37 & \cellcolor{tabfirst}36.57 \\
                              \cellcolor{gray!20} & \cellcolor{gray!20} SSIM$\uparrow$    & \cellcolor{tabfirst}0.98                     & \cellcolor{tabthird}0.97  & \cellcolor{tabfirst}0.98  & \cellcolor{tabfirst}0.98  & \cellcolor{tabfirst}0.98  & \cellcolor{tabsecond}0.98  & \cellcolor{tabthird}0.96  & \cellcolor{tabfirst}0.96  & \cellcolor{tabsecond}0.96  \\
                              \multirow{-3}{*}{Gaussian-SLAM+Ours} \cellcolor{gray!20} &\cellcolor{gray!20}LPIPS$\downarrow$  & \cellcolor{tabfirst}0.075                    & \cellcolor{tabfirst}0.07  & \cellcolor{tabfirst}0.07  & \cellcolor{tabfirst}0.08  & \cellcolor{tabfirst}0.07  & \cellcolor{tabfirst}0.07  & \cellcolor{tabfirst}0.08  & \cellcolor{tabfirst}0.06  & \cellcolor{tabsecond}0.11  \\ \hline
\end{tabular}}
\caption{Quantitative image reconstruction performance  on Replica dataset~\cite{replica}. We use the PSNR $\uparrow$, SSIM $\uparrow$, and LPIPS $\downarrow$ as the metrics and compare them with other SOTA methods.}
\label{tab:replica1}
\end{table*}

In order to further quantize the attributes stored in 3D Gaussian ellipsoids, we create additional codebooks for anchor feature. Our experiments show that 1-byte
indices allow maximum compression with minimal quality degradation of color and opacity. Hence, with this quantization strategy, we reduce the request memory for these attributes. Since N can be a number in the hundreds of thousands or millions with the SLAM system operation, our compact scene representation can significantly reduce the memory and storage usage.
\subsection{Tracking and Local-to-Global Bundle Adjustment}
\label{sec:ba}
Our tracking and bundle adjustment
are performed by minimizing our objective functions. The camera pose is initialized for a new time step by a constant velocity forward projection of the pose parameters.
 For each pixel $p$, the color and opacity of all Gaussian ellipsoids are computed and blended using this formula:
\begin{equation}
    C(p)=\sum_{i \in N} \boldsymbol{c}_{\boldsymbol{i}} f_i(p) \prod_{j=1}^{i-1}\left(1-f_j(p)\right)
\end{equation}
where $c_i$ denotes the color of Gaussian ellipsoids. We also propose a similar depth rendering formula:
\begin{equation}
    D(p)=\sum_{i=1}^n d_i f_i(\mathbf{p}) \prod_{j=1}^{i-1}\left(1-f_j(p)\right)
\end{equation}
We also render a silhouette image to determine visibility:
\begin{equation}
    S(p)=\sum_{i=1}^n f_i(p) \prod_{j=1}^{i-1}\left(1-f_j(p)\right)
\end{equation}
The color and depth loss is defined as:
\begin{equation}
\mathcal{L}_{c}=\frac{1}{N} \sum_{i=1}^N\left(\hat{\mathbf{C}}_i-\mathbf{C_i}\right)^2,\quad
\mathcal{L}_d=\frac{1}{\left|R_i\right|} \sum_{i \in R_i}\left(\mathbf{\hat{D_i}}-\mathbf{D_i}\right)^2
\end{equation}
where $R_i$ is the set of rays that have a valid depth observation.
The ICP loss is common in traditional SLAM methods based on sparse point clouds. Since 3D Gaussian is also based on a point cloud representation, we introduce this loss to improve the scene's geometric representation and consistency. We formulate ICP loss:
\begin{equation}
    \mathcal{L}_{icp} =\sum_{i=1}^n \omega_i\left  \|R\cdot p_i+t-q_i    \right \|^2  
\end{equation}
where $p_i$ denotes the 3D Gaussian point i, $q_i$ denotes the closest 
 points. 
The tracking loss is formulated as follows:
\begin{equation}
    L_{\mathrm{t}}=\sum_{\mathbf{p}}(S(\mathbf{p})>0.99)\left(\mathcal{L}_{c}+\lambda_1 \mathcal{L}_{d}+\lambda_2\mathcal{L}_{icp}\right)
\end{equation}
We use the rendered visibility silhouette to select the well-optimized pixels for camera tracking, which can improve the tracking accuracy for the new frames.

\noindent\textbf{Local-to-Global Bundle Adjustment.} 
For global consistency and accuracy, our system maintains a significantly larger global keyframe database than other GS-based SLAM systems.
We randomly sample a total number of N rays from our global keyframe database to optimize
our scene representation as well as camera poses. This phase optimizes a loss similar to tracking loss, and we also add an SSIM loss to RGB rendering. The local-to-global bundle adjustment is performed to optimize the scene representation with the camera pose.  Our local-to-global BA method can effectively reduce cumulative errors and enhance the robustness of pose estimation, especially for long sequences and large scenes.

\begin{figure}[t]
  \centering
\includegraphics[width=\linewidth]{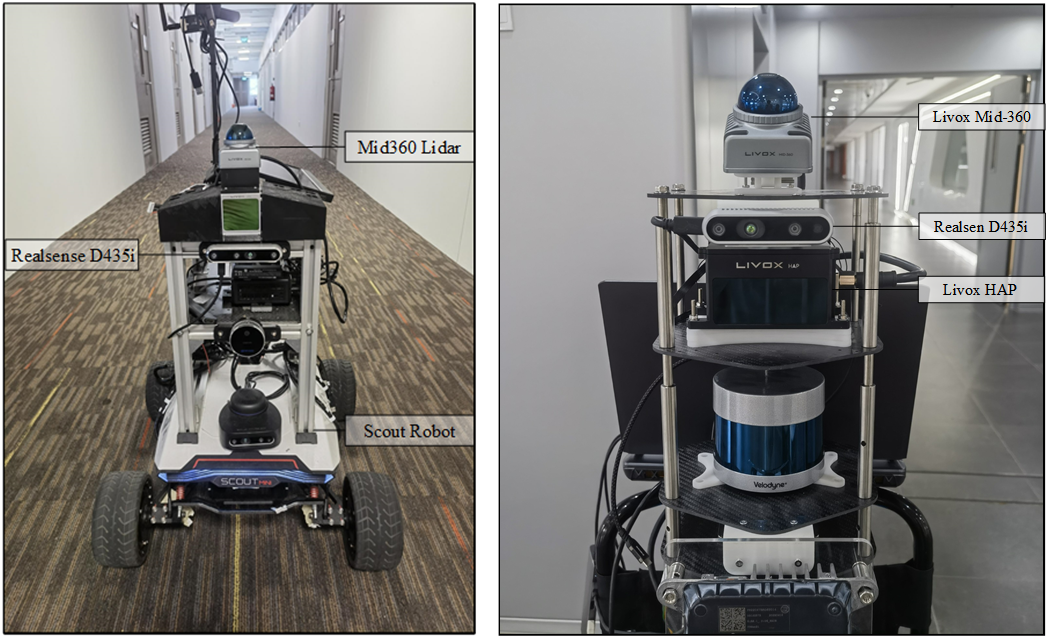}
  \caption{Visualization of the different robot platformsensors in our dataset. }
  \label{fig:real-world2}
\end{figure}

\begin{table}
\scalebox{0.8}{
\setlength{\tabcolsep}{0.7mm}{
\begin{tabular}{llccc}
\toprule
Scene ID               & \multicolumn{1}{c}{Sequence ID} & \multicolumn{1}{c}{Scale} & \multicolumn{1}{c}{Total Length} & \multicolumn{1}{c}{Total duration} \\ \midrule
Indoor\_corridor       & Corridor1,2,3,4                 & \textgreater{}1000 $m^2$       & 1482.75m                                    & 23487s                              \\
Indoor\_hall       & Hall1,2,3,4                 & \textgreater{}1000 $m^2$       & 582.59m                                    & 18987s                              \\
Indoor\_hall2       & Hall1,2,3,4                 & \textgreater{}1000 $m^2$       & 683.47m                                    & 19465s                              \\
\bottomrule
\end{tabular}}}
\caption{Summary of scene ID, and the sequence ID in each scene. We present the scales of different scenes, as well as the cumulative lengths of the datasets and the total duration (number of frames). `min." denotes minutes.}

\label{tab:scene}
\end{table}

\section{Experiments}
\textbf{Datasets and Evaluation Metrics.} We evaluate our method on various datasets: Replica~\cite{replica}, Scannet~\cite{scannet1}, and TUM-RGBD~\cite{tum}, following the settings form SplaTAM~\cite{splatam}. The replica is a synthetic dataset, and we select eight small room scene sequences for evaluation. We
select six real-world scenes from the ScanNet~\cite{scannet1} dataset for relatively large-scale indoor scenes. It was captured using the RGB-D camera of an iPad. We also select five scenes from the TUM RGB-D~\cite{tum} dataset as it is a common dataset for evaluating camera tracking accuracy. The TUM RGB-D dataset~\cite{tum} was recorded with a Kinect V1 sensor at a resolution of 640×480 @ 30Hz.
We follow the evaluation metrics from SplaTAM~\cite{splatam}. We use PSNR, SSIM, and LPIPS to measure rendering performance. For camera pose estimation, we use
the average absolute trajectory error (ATE RMSE$[cm]$ $\downarrow$). We also report the requirement of computing resources by showing the tracking and mapping time every iteration,tracking and mapping time every frame,  rendering
FPS, decoder parameter, and memory usage. 
We also collect our own dataset based on our mobile robot's embedded platform, including data from multiple sensor modalities to further validate our algorithm on a real-world platform across various indoor and outdoor scenes. 

\begin{figure}[t]
  \centering
\includegraphics[width=\linewidth]{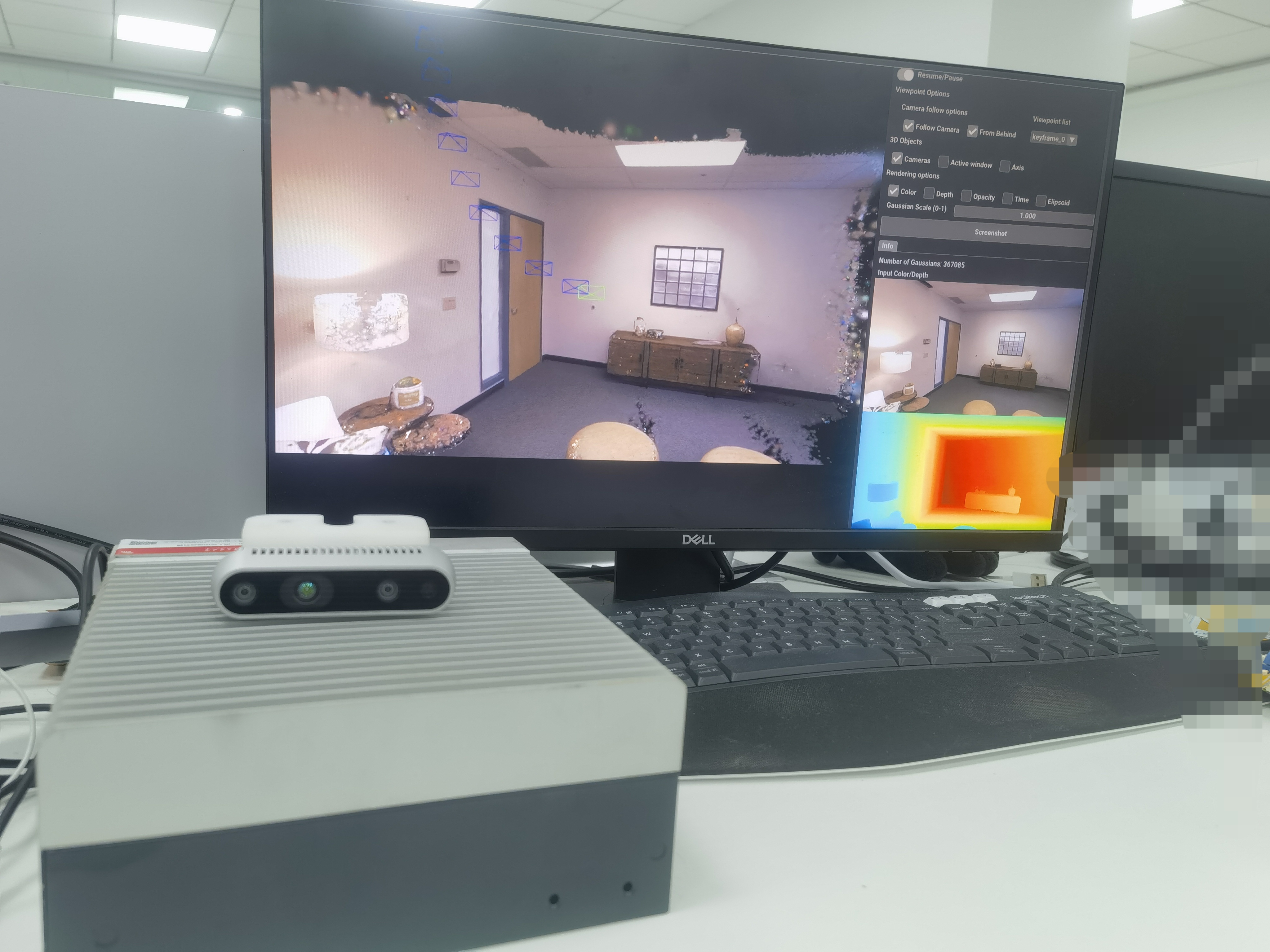}
  \caption{The proposed system has real-time performance on
embedded platforms, such as Jetson Xavier MIC-730AI Developer Kit.}
  \label{fig:real-world}
\end{figure}

\noindent\textbf{Implementation Details.} Our method is implemented in Python using the PyTorch framework, incorporating CUDA for Gaussian Splatting, and trained on a desktop PC with NVIDIA RTX 3090Ti GPU. We further tested our method on a laptop and a Jetson Xavier Developer Kit. The laptop is
equipped with an NVIDIA RTX 3080ti 16 GB Laptop GPU,
an Intel Core i9-12900HX, and 32 GB RAM. We extended the existing differentiable Gaussian Splatting rasterization code by adding functions to manage depth, pose, and cumulative opacity during both forward and backward propagation.

\begin{table*}[]
\centering
\scalebox{0.90}{
\setlength{\tabcolsep}{1.0mm}{
\begin{tabular}{llccccccccc}
\toprule
Methods                                                                 & Metrics & Avgerage & Room0   & Room1   & Room2   & Office0  & Office1  & Office2  & Office3  & Office4    \\ \hline
\multicolumn{11}{l}{\cellcolor[HTML]{EEEEEE}{\textit{NeRF-based Methods}}} \\
\multirow{2}{*}{NICE-SLAM\cite{niceslam}}                                                 & Depth $\downarrow$  & 2.99 & 1.83  & 1.46  & 2.03  & 1.36  & 1.74  & 8.31  & 4.95  & 2.03   \\ & F1[\%] $\uparrow$ & 44.5 & 45.5 & 44.6 & 43.9 & 50.5 & 52.5 & 40.5 & 40.3 & 36.2 \\ 
\multirow{2}{*}{Vox-Fusion\cite{voxfusion}}                                                 & Depth $\downarrow$  & 2.49 & 1.13  & 1.94  & 2.12  & 2.30  & 3.34  & 4.08  & 2.92  & 1.57   \\ & F1[\%] $\uparrow$ & 52.9 & 70.5 & 35.3 & 60.4 & 50.5 & 52.5 & 40.5 & 40.3 & 36.2 \\ 
\multirow{2}{*}{ESLAM\cite{eslam}}                                                 & Depth $\downarrow$  & 1.18 & 0.97  & 1.07  & 1.28  & 0.86  & 1.26  & 1.71  & 1.43  & 1.06    \\ & F1[\%] $\uparrow$ & 79.1 & 81.4 & 82.3 & 83.5 & 78.7 & 75.4 & 77.2 & 75.3 & 79.0 \\ 
\multirow{2}{*}{Co-SLAM\cite{coslam}}                                               & Depth $\downarrow$   & 1.51 & 1.05  & 0.85  & 2.37  & 1.24  & 1.48  & 1.86  & 1.66  & 1.54    \\  & F1[\%] $\uparrow$ & 77.9 & 79.8 & 81.4 & 82.3 & 77.3 & 74.7 & 75.9 & 74.1 & 77.8  \\ 
                              
\multirow{2}{*}{GO-SLAM\cite{goslam}}                                               & Depth $\downarrow$  & 2.56  & 1.57  & 2.43  & 1.47  & 1.63  & 2.31  & 1.71  & 1.47  & 4.68  \\  & F1[\%] $\uparrow$ & 70.9 & 72.8 & 74.4 & 75.1 & 71.8 & 68.7 & 69.3 & 68.3.1 & 70.2  \\ 
                                 
\multirow{2}{*}{Point-SLAM\cite{pointslam}} & Depth $\downarrow$  & \cellcolor{tabfirst}0.53  & \cellcolor{tabfirst}0.22  & 0.46  & \cellcolor{tabfirst}0.30  & 0.57  & \cellcolor{tabfirst}0.49  & \cellcolor{tabfirst}0.51  & \cellcolor{tabfirst}0.46  & \cellcolor{tabthird}0.44  \\ & F1 [\%] $\uparrow$ & 89.5  & 86.4  & \cellcolor{tabfirst}92.0  & \cellcolor{tabthird}90.5 & 91.4 & 89.6 & \cellcolor{tabfirst}89.5  & \cellcolor{tabfirst}88.6 & \cellcolor{tabthird}85.9 \\   \hdashline
                                                                       \multicolumn{11}{l}{\cellcolor[HTML]{EEEEEE}{\textit{3DGS-based Methods}}} \\
\multirow{2}{*}{SplaTAM\cite{splatam}}                                               & Depth $\downarrow$ & 0.72 & \cellcolor{tabthird}0.43  & 0.38  & 0.54  & \cellcolor{tabthird}0.44  & 0.66  & 1.05  & \cellcolor{tabthird}1.60  & 0.68    \\ & F1 [\%] $\uparrow$ & 86.3  & 89.5  & 88.5  & 88.4 & 91.8 & 89.8 & 84.8  & 77.9 & 80.8 \\  
                                               SplaTAM+Ours \cellcolor{gray!20} & \cellcolor{gray!20}Depth $\downarrow$ & 0.70 & \cellcolor{tabsecond}0.42  & \cellcolor{tabthird}0.37  & 0.53  & \cellcolor{tabthird}0.44  & 0.64  & 1.02  & \cellcolor{tabsecond}1.57  & 0.66    \\  \cellcolor{gray!20}
                                               & \cellcolor{gray!20}F1 [\%] $\uparrow$ & 86.9  & 89.8  & 88.9  & 88.9 & \cellcolor{tabthird}92.0 & \cellcolor{tabsecond}90.8 & 85.3  & 78.4 & 81.3 \\ 
\multirow{2}{*}{MonoGS~\cite{monogs}}                                               & Depth $\downarrow$  & 0.73 & 0.45  & 0.39  & \cellcolor{tabsecond}0.52  & \cellcolor{tabfirst}0.43  & 0.69  & 1.08  & 1.63  & 0.69    \\ & F1 [\%] $\uparrow$ & 86.6  & 89.1  & 88.7  & 89.1 & 91.5 & 89.4 & 85.5  & 78.3 & 81.5 \\                         MonoGS+Ours       \cellcolor{gray!20}               & \cellcolor{gray!20}Depth $\downarrow$  & 0.70 & 0.44  & 0.38  & 0.53  & \cellcolor{tabfirst}0.43  & 0.70  & 1.05  & 1.61  & 0.68    \\ \cellcolor{gray!20} & \cellcolor{gray!20} F1 [\%] $\uparrow$ & 87.1  & \cellcolor{tabthird}89.3  & 88.9  & 89.1 & 91.3 & 89.5 & 85.8  & 78.5 & 81.8 \\
\multirow{2}{*}{Gaussian-SLAM~\cite{gaussianslam}}                                               & Depth $\downarrow$ & \cellcolor{tabthird}0.68  & 0.61  & \cellcolor{tabfirst}0.25  & 0.54  & 0.48 & \cellcolor{tabthird}0.51  & \cellcolor{tabthird}0.97  & 1.63  & \cellcolor{tabsecond}0.42  \\ & F1 [\%] $\uparrow$ & \cellcolor{tabthird}89.1  & \cellcolor{tabsecond}89.5  & \cellcolor{tabthird}89.7  & \cellcolor{tabsecond}90.9 & \cellcolor{tabfirst}92.1 & \cellcolor{tabthird}90.5 & \cellcolor{tabthird}87.8  & \cellcolor{tabthird}84.7 & \cellcolor{tabsecond}87.6 \\ 
Gaussian-SLAM+Ours   \cellcolor{gray!20}                                          & \cellcolor{gray!20}Depth $\downarrow$ & \cellcolor{tabsecond}0.66  & 0.60  & \cellcolor{tabsecond}0.26  & \cellcolor{tabsecond}0.52  & \cellcolor{tabthird}0.44  & \cellcolor{tabsecond}0.50  & \cellcolor{tabsecond}0.92  & \cellcolor{tabthird}1.60  & \cellcolor{tabfirst}0.40  \\ \cellcolor{gray!20} & \cellcolor{gray!20} F1 [\%] $\uparrow$ & \cellcolor{tabfirst}89.6  & \cellcolor{tabfirst}89.9  & \cellcolor{tabsecond}91.1  & \cellcolor{tabfirst}91.2 & \cellcolor{tabfirst}92.1 & \cellcolor{tabfirst}91.1 & \cellcolor{tabsecond}88.1  & \cellcolor{tabsecond}85.2 & \cellcolor{tabfirst}87.9 \\ \hline
\end{tabular}}}
\caption{Scene Reconstruction Performance on Replica dataset~\cite{replica}. We use Depth L1 (cm), F1 score as the metrics. We select 8 sequences following the settings from previous methods. }
\label{tab:render2}
\end{table*}

\begin{table*}[]
\vspace{-0.2cm}
\centering
\scalebox{0.99}{
\setlength{\tabcolsep}{1.6mm}{
\begin{tabular}{lcccccc}
\toprule
\multirow{2}{*}{Methods} & \multicolumn{3}{c}{Training View}                                               & \multicolumn{3}{c}{Novel View}                                                  \\
                         & \multicolumn{1}{l}{PNSR$\uparrow$} & \multicolumn{1}{l}{SSIM$\uparrow$} & \multicolumn{1}{l}{LPIPS$\downarrow$} & \multicolumn{1}{l}{PNSR$\uparrow$} & \multicolumn{1}{l}{SSIM$\uparrow$} & \multicolumn{1}{l}{LPIPS$\downarrow$} \\ \hline
                         NICE-SLAM~\cite{niceslam}                  & 13.9                     & 0.47                     & 0.81                      & 14.4                    & 0.39                     & 0.88                      \\
Co-SLAM~\cite{coslam}                  & 14.7                     & 0.49                     & 0.79                      & 14.2                    & 0.39                     & 0.89                      \\
ESLAM~\cite{eslam}                    & 25.8                     & 0.84                     & 0.51                      & 19.8                    & 0.79                     & 0.50                      \\ Photo-SLAM~\cite{photoslam}                  & \cellcolor{tabthird}28.3                     & 0.86                     & 0.43                      & 25.7                        & 0.84                       & 0.44                         \\
SplaTAM~\cite{splatam}                    & 27.8                     & 0.86                     & 0.44                      & 25.4                    & 0.85                     & 0.45   \\
\cellcolor{gray!20}SplaTAM+Ours                    & 28.1                     & \cellcolor{tabsecond}0.88                     & 0.43                      & 25.7                    & \cellcolor{tabsecond}0.86                     & 0.44    \\ MonoGS~\cite{monogs}                    & 28.1                     & 0.87                     & 0.43                      & 25.8                    & \cellcolor{tabsecond}0.86                     & \cellcolor{tabthird}0.43  \\ \cellcolor{gray!20}MonoGS+Ours                    & \cellcolor{tabthird}28.3                     & \cellcolor{tabsecond}0.88                     & \cellcolor{tabthird}0.42                      & \cellcolor{tabsecond}26.1                    & \cellcolor{tabfirst}0.87                     & \cellcolor{tabthird}0.43  \\ Gaussian-SLAM~\cite{gaussianslam}                    & \cellcolor{tabsecond}28.7                     & \cellcolor{tabsecond}0.88                     &\cellcolor{tabfirst} 0.40                      & \cellcolor{tabthird}26.0                    & 0.85                     & \cellcolor{tabfirst}0.41  \\ \cellcolor{gray!20}Gaussian-SLAM+Ours                   & \cellcolor{tabfirst}28.9                     & \cellcolor{tabfirst}0.89                     & \cellcolor{tabfirst}0.40                      & \cellcolor{tabfirst}26.3                    & \cellcolor{tabsecond}0.86                     & \cellcolor{tabfirst}0.41  \\ 
\bottomrule
\end{tabular}}}
\caption{Quantitative rendering results of our proposed framework with existing SLAM systems on our dataset in long corridor sequence.}
\label{tab:ourdataset}
\end{table*}

\begin{table*}[t]
\centering

\scalebox{0.78}{
\setlength{\tabcolsep}{0.8mm}{
\begin{tabular}{l|lccccccc}
\hline
                         & Methods      & Track/Iter. $\downarrow$ & Map/Iter. $\downarrow$ & Track/Frame $\downarrow$& Map/Frame $\downarrow$& Render FPS $\uparrow$& Decoder Param. $\downarrow$& Memory $\downarrow$  \\ \hline
\multirow{10}{*}{\rotatebox{90}{Replica\cite{replica} }} & NICE-SLAM \cite{niceslam}    & 6.98ms    & 28.88ms & 68.54ms   & 1.23s   & 0.30       & 0.06M      & 48.48MB  \\
                         & ESLAM \cite{eslam}        & 6.85ms    & 19.98ms & 54.80ms   & 0.29s   & 2.82       & 0.003M     & 27.12MB  \\
                         & Co-SLAM \cite{coslam}      & 6.38ms    & 14.25ms & 63.93ms   & 0.15s   & 3.68       & 0.013M     & 24.85MB  \\
                         & Point-SLAM \cite{pointslam}   & 0.57ms    & 34.85ms & 22.72ms   & 10.47s  & 1.33       & 0.127M     & 55.42MB  \\ \cdashline{2-9}
                         & SplaTAM \cite{splatam}      & 24.23ms   & 22.83ms & 2.18s     & 1.37s   & 175.64     & \cellcolor{tabfirst}0M         & 273.09MB \\
                         & \cellcolor{gray!20}SplaTAM+Ours & 16.83ms   & \cellcolor{tabthird}15.71ms & 1.51s     & \cellcolor{tabfirst}0.94s   & 398.45     & \cellcolor{tabfirst}0M         & 117.36MB \\
                         & MonoGS \cite{monogs}       & 15.39ms   & \cellcolor{tabsecond}12.49ms & 1.54s     & 1.88s   & 317.45     & \cellcolor{tabfirst}0M         & \cellcolor{tabsecond}84.41MB \\
                         & \cellcolor{gray!20}MonoGS+Ours  & \cellcolor{tabfirst}13.25ms   & \cellcolor{tabfirst}10.74ms & 1.32s     & \cellcolor{tabsecond}1.61s   & \cellcolor{tabsecond}447.29     & \cellcolor{tabfirst}0M         & \cellcolor{tabfirst}73.31MB  \\
                         & Gaussian-SLAM \cite{gaussianslam}       & 14.09ms   & 20.85ms & \cellcolor{tabfirst}0.79s     & 2.08s   & 321.39     & \cellcolor{tabfirst}0M         & 101.48MB \\
                         & \cellcolor{gray!20}Gaussian-SLAM+Ours  & \cellcolor{tabsecond} 13.65ms   & 17.97ms & \cellcolor{tabsecond}0.81s     & \cellcolor{tabthird}1.79s   & \cellcolor{tabfirst}458.53     & \cellcolor{tabfirst}0M         & \cellcolor{tabthird}96.03MB  \\
                         & \cellcolor{gray!20}Ours(Jetson)  & 14.49ms   & 18.15ms & 0.86s     & 1.81s   & 439.23     & \cellcolor{tabfirst}0M         & \cellcolor{tabthird}96.03MB  \\
                         & \cellcolor{gray!20}Ours(Laptop)  & \cellcolor{tabthird}13.88ms   & 18.01ms & \cellcolor{tabthird}0.83s     & 1.80s   & \cellcolor{tabthird}445.53     & \cellcolor{tabfirst}0M         & \cellcolor{tabthird}96.03MB  \\
                         \hline
\multirow{12}{*}{\rotatebox{90}{Scannet\cite{scannet1}}} & NICE-SLAM \cite{niceslam}    & 12.3ms    & 125.3ms & 0.62s     & 7.52s   & 0.27       & 0.07M      & 84.25MB  \\
                         & ESLAM \cite{eslam}       & 7.54ms    & 23.4ms  & 0.23s     & 0.70s   & 2.74       & 0.004M     & 67.95MB  \\
                         & Co-SLAM \cite{coslam}      & 7.81ms    & 20.6ms  & 78ms      & 0.20s   & 3.41       & 0.015M     & 48.81MB  \\
                         & Point-SLAM \cite{pointslam}   & 0.63ms    & 42.29ms & 64ms      & 12.69s  & 1.28       & 0.131M     & 57.21MB  \\ \cdashline{2-9}
                         & SplaTAM \cite{splatam}     & 28.32ms   & 26.37ms & 2.55s     & \cellcolor{tabsecond}1.58s   & 155.69     & \cellcolor{tabfirst}0M         & 178.09MB \\
                         & \cellcolor{gray!20}SplaTAM+Ours & 20.85ms   & 19.25ms & \cellcolor{tabthird}1.88s     & \cellcolor{tabfirst}1.23s   & 387.93     & \cellcolor{tabfirst}0M         & \cellcolor{tabsecond}92.48MB  \\
                         & MonoGS \cite{monogs}       & 18.21ms   & \cellcolor{tabthird}16.91ms & \cellcolor{tabsecond}1.82s     & 2.54s   & 301.39     & \cellcolor{tabfirst}0M         & 117.47MB \\
                         & \cellcolor{gray!20}MonoGS+Ours  & \cellcolor{tabfirst}15.21ms   & \cellcolor{tabfirst}13.27ms & \cellcolor{tabfirst}1.52s     & 1.98s   & \cellcolor{tabsecond}418.43     & \cellcolor{tabfirst}0M         & \cellcolor{tabfirst}74.35MB  \\ 
                         & Gaussian-SLAM \cite{gaussianslam}       & 16.49ms   & 16.98ms & 3.29s     & 1.99s   & 314.39     & \cellcolor{tabfirst}0M         & 127.43MB \\
                         & \cellcolor{gray!20}Gaussian-SLAM+Ours  & \cellcolor{tabsecond}15.33ms   & \cellcolor{tabsecond}16.34ms & 1.53s     & \cellcolor{tabthird}1.63s   & \cellcolor{tabfirst}423.18     & \cellcolor{tabfirst}0M         & \cellcolor{tabthird}95.79MB  \\
                         &\cellcolor{gray!20}Ours(Jetson)      & \cellcolor{tabthird}16.19ms   & 16.76ms & 3.23s     & 1.68s   & 401.25     & \cellcolor{tabfirst}0M         & \cellcolor{tabthird}95.79MB \\
                         &\cellcolor{gray!20}Ours(Laptop)       & 15.98ms   & 16.65ms & 3.19s     & 1.66s   & \cellcolor{tabthird}409.43     & \cellcolor{tabfirst}0M         & \cellcolor{tabthird}95.79MB \\ \hline
\end{tabular}}}
\caption{Runtime and memory performance evaluation on Replica~\cite{replica} and ScanNet~\cite{scannet1} dataset. The Decoder Param denotes the parameter number of MLPs. The memory is the memory usage of the checkpoint.}
\label{tab:runtime}
\end{table*}

\begin{figure*}[t]
  \centering
\includegraphics[width=\linewidth]{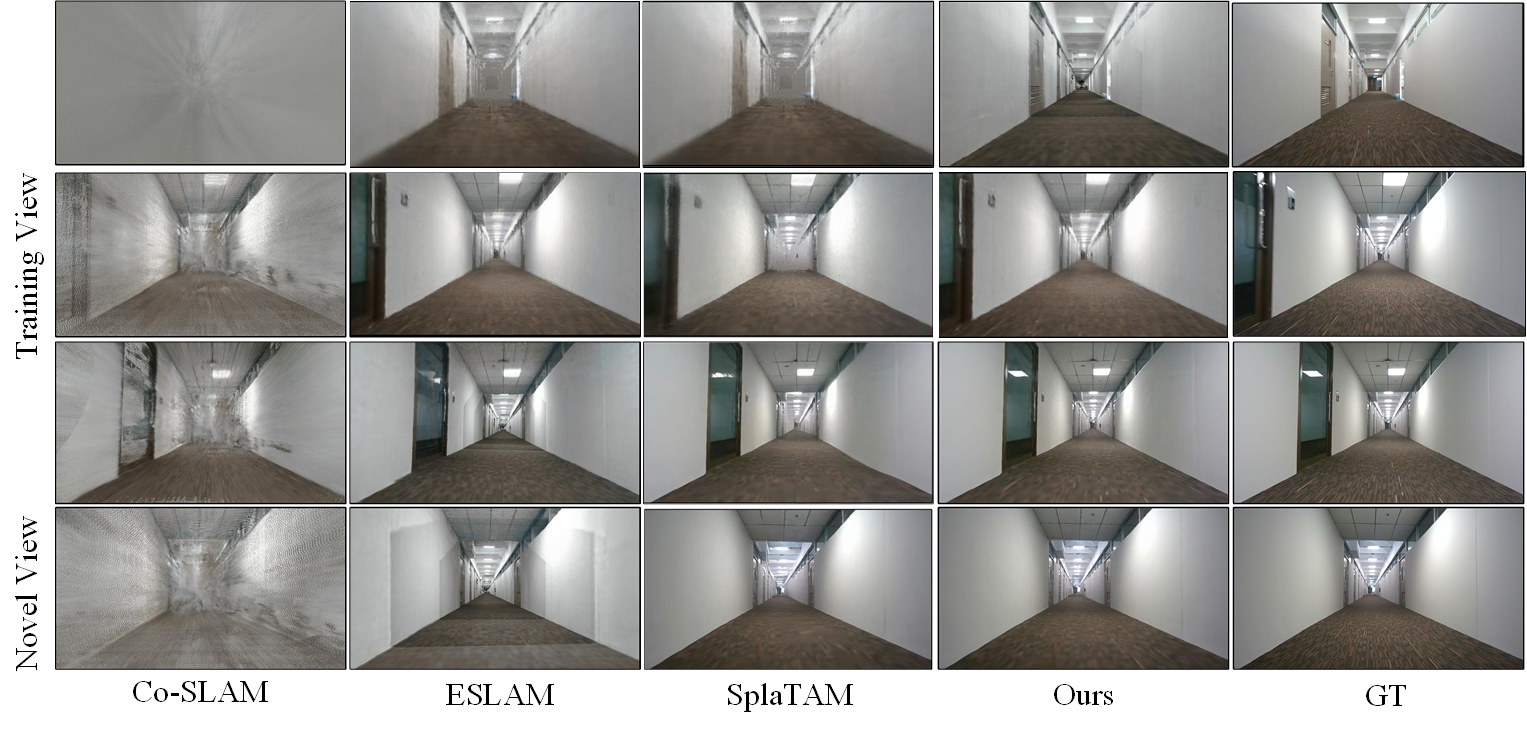}
  \caption{Qualitative comparison of our proposed method’s rendering performance with existing SOTA methods on our dataset in long corridor scene. }
  \label{fig:ourdataset2}
\end{figure*}

\begin{figure*}[t]
  \centering
\includegraphics[width=\linewidth]{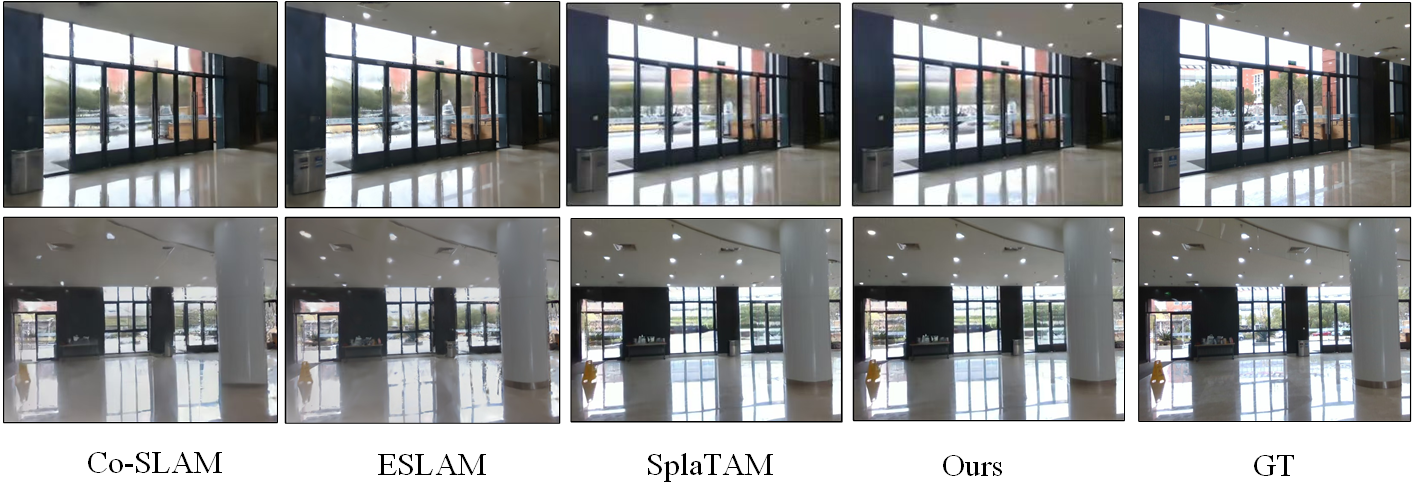}
  \caption{Qualitative comparison of our proposed method’s rendering performance with existing SOTA methods on our dataset in hall scene. }
  \label{fig:ourdataset3}
\end{figure*}

\begin{figure*}[t]
  \centering
\includegraphics[width=0.95\linewidth]{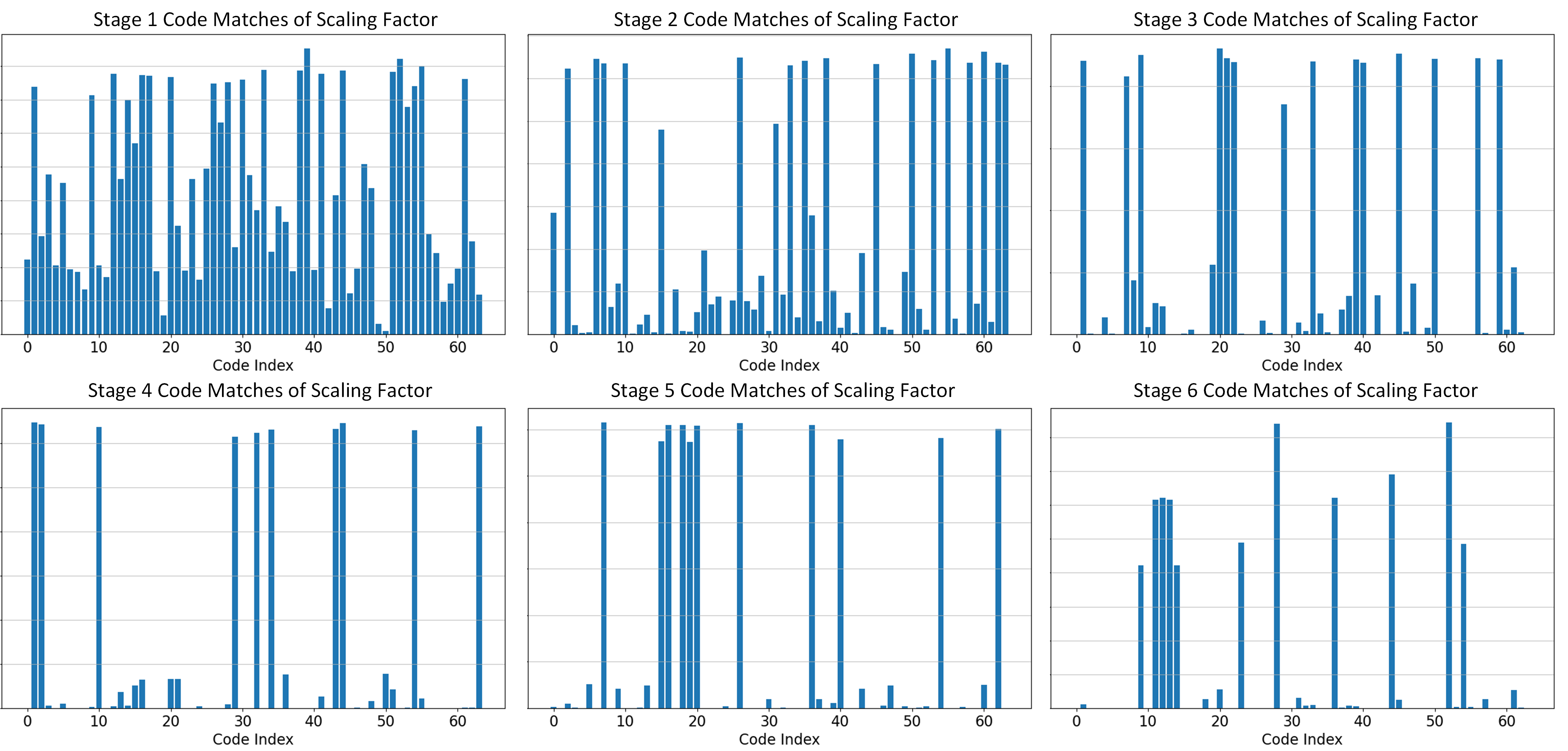}
  \caption{We visualized the learned codebook indices for scaling factor vector in the replica dataset~\cite{replica}. We can observe that as the stage progresses, the distribution of the codebook becomes increasingly compact. We visualize all the stages of our codebook.}
  \label{fig:rvq2}
\end{figure*}

\begin{table*}[t]
\centering
\scalebox{0.8}{
\setlength{\tabcolsep}{0.8mm}{
\begin{tabular}{l|ccccccc}
\hline
\multirow{2}{*}{Method} & \multicolumn{4}{c}{Accuracy} & \multicolumn{3}{c}{Real-time performance and Memory} \\  
                        & PSNR $\uparrow$   & SSIM $\uparrow$ & LPIPS $\downarrow$  & ATE RMSE $\downarrow$  & Training Time $\downarrow$         & Render FPS$\uparrow$             & Memory$\downarrow$            \\ \midrule
                        w/o voxel                & 33.46  & 0.95 & 0.11  & \cellcolor{tabthird}0.28 & 1.89H          & 235.37          & 215.34MB \\
w/o mask                & 33.53  & 0.96 & 0.11  & \cellcolor{tabthird}0.28 & 1.90H          & 245.59          & 216.09MB \\
w/o rvq                 & \cellcolor{tabsecond}33.57  & \cellcolor{tabsecond}0.97 & \cellcolor{tabsecond}0.10  & \cellcolor{tabfirst}0.27 & \cellcolor{tabthird}1.47H          & \cellcolor{tabthird}274.26          & \cellcolor{tabthird}167.81MB          \\
w/o icp loss           & \cellcolor{tabthird}33.90  & \cellcolor{tabsecond}0.97 & \cellcolor{tabsecond}0.10  & 0.30 & \cellcolor{tabfirst}1.37H          & \cellcolor{tabfirst}398.45          & \cellcolor{tabfirst}117.36MB          \\
SplaTAM+Ours                   & \cellcolor{tabfirst}34.31  & \cellcolor{tabfirst}0.98 & \cellcolor{tabfirst}0.09  & \cellcolor{tabfirst}0.27 & \cellcolor{tabfirst}1.36H          & \cellcolor{tabfirst}398.45         & \cellcolor{tabfirst}117.36MB          \\ \bottomrule
\end{tabular}}}
\caption{The ablation study on Replica~\cite{replica} dataset. We conduct experiments to verify the effectiveness of our method. Our full
model achieves better pose estimation performance as well as faster training and rendering speed and lower memory usage.'H' represents hours.}
\label{tab:ablation}
\end{table*}

    \begin{table*}[h!]
\centering
\scalebox{0.8}{
\setlength{\tabcolsep}{0.8mm}{
\begin{tabular}{l|ccccccc}
\hline
\multirow{2}{*}{Method} & \multicolumn{4}{c}{Accuracy} & \multicolumn{3}{c}{Real-time performance and Memory} \\  
                        & PSNR $\uparrow$   & SSIM $\uparrow$ & LPIPS $\downarrow$  & ATE RMSE $\downarrow$  & Training Time $\downarrow$         & Render FPS$\uparrow$             & Memory$\downarrow$            \\ \midrule
                        Step L=4                & 32.49  & 0.97 & 0.09  &   0.28 & 1.34H          & 396.48           & 121.49MB \\
Step L=10                 &   33.83  &   0.98 &   0.09  &  0.27 &   1.40H         &  392.34           &   138.37MB          \\
Codebook C= 48          &   32.88  &   0.97 &   0.10  & 0.27 &  1.35H          &  398.69          &  122.04MB          \\
Codebook C= 80          &   33.94  &   0.98 &   0.09  & 0.27 &  1.40H          &  397.21          &  126.43MB          \\
window size=30                & 31.68  & 0.95 & 0.11  &   0.31 & 1.35H          & 400.27           & 119.32MB \\
window size=20                 &   33.89  &   0.98 &   0.09  &  0.28 &   1.39H         &  392.34           &   131.14MB          \\
Ours(window size=24,L=4,C=64)                   &  34.31  &  0.98 &  0.09  &  0.27 &  1.36H          &  398.45          &  117.36MB          \\ \bottomrule
\end{tabular}}}

\caption{The ablation study on Replica~\cite{replica} dataset. We conduct experiments to verify the effectiveness of our method. }
\label{tab:ablation3}
\end{table*}

\begin{table}[]
\centering
\scalebox{0.8}{
\setlength{\tabcolsep}{0.8mm}{
\begin{tabular}{lcccc}
\toprule
                          & Sil. Thresh. & ATE    & Depth L1 & PSNR  \\ \midrule
w/o velo. prop.           & 0.99         & 2.75   & 2.05     & 25.57 \\
w/o sil. mask             & 0.99         & 115.18 & 0.49     & 14.15 \\ 
w/o track. and map. color & 0.99           & 88.23  & -        & -     \\
w/o track. and map. depth & 0.99         & 1.38   & 12.87    & 31.05 \\
Ours                      & \textbf{0.99}         & \textbf{0.24}   & \textbf{0.42}     & \textbf{32.99}  \\ \bottomrule
\end{tabular}}}
\caption{Some extended ablation experiments of camera tracking and color and depth loss on Replica~\cite{replica} room0 scene.}
\label{tab:ablation2}
\end{table}
\noindent\textbf{Baselines.}
The main baseline methods we compare to are SplaTAM~\cite{splatam}, MonoGS~\cite{monogs}, Gaussian-SLAM~\cite{gaussianslam} as the representative of the GS-based SLAM systems. Our method is also a plug-and-play method, which can improve the speed and memory usage for all GS-based methods. We also compared to other NeRF-based SLAM methods, such as NICE-SLAM~\cite{niceslam}, Co-SLAM~\cite{coslam}, ESLAM~\cite{eslam}, Vox-Fusion~\cite{voxfusion}, Point-SLAM~\cite{pointslam}, GO-SLAM~\cite{goslam}, PLG-SLAM~\cite{plgslam}, Loopy-SLAM~\cite{loopyslam}. On the TUM-RGBD dataset, we also compare
against three traditional SLAM systems: Kintinuous~\cite{Kintinuous},
ElasticFusion~\cite{elasticfusion}, and ORB-SLAM2~\cite{orbslam2}. 
\subsection{Dataset Collection}
We introduce a large-scale Neural SLAM dataset that spans a diverse range of scenarios, from small indoor rooms to expansive hall environments. The dataset provide high-quality trajectory ground truth.  We believe this dataset has the potential to significantly advance research and innovation within the SLAM community.

Our collected dataset includes information from multiple sensor modalities, including various types of Livox HAP LiDAR, Livox Mid360, IMU, and RealSense RGB-D camera.  In Fig. ~\ref{fig:real-world2}, we visualize our data collection equipment, utilizing two different sensor modules for data acquisition. The specific sensor details are labeled in the figures. In Tab.~\ref{tab:sensor}, we present the summary of sensing modalities, handware units, ROS topics, and the nominal rates on each platform. All these data have also been directly extracted from the rosbag and saved as individual files.

\subsection{Experimental Results}
In this section, we present our experimental results on three different datasets. We evaluate the camera pose estimation, the 3D Gaussian reconstruction, and the real-time performance and memory usage of different SLAM systems. 

\noindent\textbf{Camera Tracking Results.} In Tab. ~\ref{tab:replica}, Tab.~\ref{tab:scannet}, Tab.~\ref{tab:tum}, we compare our method with other SOTA methods on camera pose estimation on different datasets. Best results are highlighted as \colorbox{tabfirst}{first}, \colorbox{tabsecond}{second}, and \colorbox{tabthird}{third}.
On the synthetic dataset Replica~\cite{replica}, we can see that we successfully reduce the trajectory error over the GS-based SLAM system and achieve more accurate and robust pose estimation, shown in Tab. \ref{tab:replica}. This is thanks to our improved tracking method, which additionally uses ICP loss to enhance the association of 3D Gaussian ellipsoids.  The ScanNet dataset is a real-world dataset that has poor depth sensor information with high motion blur on RGB images. So, relatively, the implicit-based dense SLAM method has an inherent disadvantage in the ScanNet scenes. In Tab.~\ref{tab:scannet}, our method performs better than the GS-based SLAM system and similarly to the previous NeRF-based SLAM systems. In Tab.~\ref{tab:tum}, we present the experimental results on TUM RGB-D datasets. We can see that our approach still outperforms other GS-based or NeRF-based methods, and the trajectory error has nearly \textbf{10\%} reduction.
Our experiments demonstrate the effectiveness of our proposed icp loss, which can still improve the pose estimation accuracy with a number of Gaussian ellipsoids removed.

\noindent\textbf{Gaussian Reconstruction Results.} In Tab. \ref{tab:replica1}, we show the rendering quality of the Replica dataset in 8 scenes. We present the quantitative results on Replica dataset in Fig. \ref{fig:replica}, and TUM RGB-D dataset in Fig.~\ref{fig:tum}. Our method achieves similar PSNR, SSIM, and LPIPS results as SplaTAM~\cite{splatam}. Our approach achieves much better results than the NeRF-based baselines, such as Vox-Fusion~\cite{voxfusion}, NICE-SLAM~\cite{niceslam}, Co-SLAM~\cite{coslam}, and ESLAM~\cite{eslam}. In Fig.~\ref{fig:ourdataset2} and Fig.~\ref{fig:ourdataset3}, we present the scene reconstruction results on our own dataset. Additionally, in Tab.~\ref{tab:ourdataset}, we present a quantitative evaluation of scene reconstruction. The results demonstrate that our method improves reconstruction accuracy. Overall, our method can achieve better performance than the SOTA methods, while the speed and memory usage are significantly lower than other methods. Although we used only a small number of Gaussian ellipsoids and compressed the global attribute information, the overall rendering accuracy remains high. At the same time, the new compact scene representation significantly improves memory usage and rendering speed.

\noindent\textbf{Real-time Performance and Memory Usage.}
Tab. \ref{tab:runtime} illustrates the runtime performance and memory usage of our method and other GS-based and NeRF-based SLAM systems on the Replica~\cite{replica} room 0 scene and the ScanNet~\cite{scannet1} 0000 scene. We follow the settings of SplaTAM for `SplaTAM + Ours' and use 90,60 iterations per frame for tracking and mapping to get similar performance in~\cite{splatam}. We use 100, 150 iterations for tracking and mapping, following the setting of MonoGS for `MonoGS + Ours'. We use 60, 100 iterations for tracking and mapping, following the setting of Gaussian-SLAM for `Gaussian-SLAM + Ours' in Replica dataset and 200, 100 iterations in ScanNet dataset. We evaluate the time consumption of tracking and mapping every iteration and every frame. Each iteration of our system renders a full 1200$\times$980 pixel image (in Replica dataset) and 640$\times$480 pixel image (in ScanNet dataset). Compared with the GS-based SLAM system, we significantly improve the training speed (\textbf{31\%} faster), which is crucial for online SLAM systems, thanks to our compact 3D Gaussian scene presentation method. We also evaluate the rendering speed, the decoder parameters, and the memory of SOTA methods. Compared with the NeRF-based SLAM system, we achieve 447.29 FPS rendering speed on the replica dataset, which is \textbf{100$\times$} significantly faster than these methods. Note that we do not use any neural network decoder in our system, which results
in zero learnable parameters of the decoder. In Fig.~\ref{fig:real-world}, we visualize the real-time performance on embedded platform.

\noindent\textbf{Plug-and-Play on Existing Works} It is extremely encouraged that our compact scene representation can serve as a plug-and-play module, improving the memory and storage efficiency of recent SOTA works\cite{splatam,gaussianslam}. In Tab.~\ref{tab:runtime}, we incorporate our compact scene representation with their SLAM systems. As shown in Tab.~\ref{tab:runtime}, our method achieves \textbf{226$\%$} increase in rendering speed and over \textbf{2.21$\times$} reduction in memory usage, compared with SplaTAM~\cite{splatam}. Similarly, introducing our compact scene representation in MonoGS~\cite{monogs} can also improve their efficiency, nearly 141$\%$ increase in rendering speed and 1.50$\times$ reduction in memory usage.

\subsection{Ablation Study}
In this section, we conduct various experiments to verify the
effectiveness of our method. Tab. \ref{tab:ablation} illustrates a quantitative
evaluation with different settings on Replica~\cite{replica} room0 scene.

\noindent\textbf{Voxel-based Scene Representation}
In the ablation study, we evaluated the effectiveness of voxelized scene representation in Tab.~\ref{tab:ablation}. The results show that this representation not only improves scene reconstruction accuracy but also significantly enhances representation efficiency, greatly boosting rendering speed and reducing memory consumption.

\noindent\textbf{Sliding Window Mask} As shown in Tab.~\ref{tab:ablation}, the proposed sliding window mask effectively reduces the number of 3D Gaussian ellipsoids while retaining the image reconstruction performance. This demonstrates that we successfully remove the redundant and unessential 3D Gaussian ellipsoids created along with the SLAM system operation. We also conducted an ablation study on the parameters of the sliding window mask. The gaussian mask is continuously optimized in the sliding window in Tab.~\ref{tab:ablation3}. Due to our loss design, the system tends to mask out an increasing number of Gaussians. When the window size is large, this leads to many useful Gaussians being masked, resulting in a drop in accuracy. Conversely, when the window size is too small, memory consumption becomes excessive. Moreover, an excessive number of Gaussians is also unfavorable for the overall optimization. In our architectural design, we reset the accumulated gradients and masks whenever a new sliding window begins, allowing the system to relearn and thereby preventing the unbounded growth of masked Gaussians. Our mask strategy shows \textbf{50\%} increase in the storage efficiency and a \textbf{26\%} increase in rendering speed.

\noindent\textbf{Residual Vector Codebook} Our proposed codebook method achieves a reduction in memory usage while maintaining the image reconstruction performance. We also present the scaling factor codebook in different stages in Fig.~\ref{fig:rvq2}. As the stages progress, the magnitude of the codes diminishes, demonstrating that the residuals for each stage are being effectively trained to capture the geometry accurately. We have supplemented our work with an ablation study on the size of the codebook and the number of stages to investigate their impact on performance in Tab.~\ref{tab:ablation3}. As the number of steps corresponds to residual quantization (i.e., progressively quantizing residuals), increasing the number of steps requires storing more levels of residual information, which leads to a increase in memory consumption and slower inference speed.

In terms of accuracy, a larger number of steps yields only marginal improvements in image reconstruction quality. Regarding the codebook size, increasing it leads to a slight improvement in reconstruction accuracy due to the higher resolution. However, many codebook indices remain unused, which increases memory consumption and slows down training. On the other hand, reducing the codebook size results in a drop in accuracy.

\noindent\textbf{ICP Loss} In Tab.~\ref{tab:ablation}, we can see that the proposed icp loss effectively improves the accuracy of camera pose estimation. As the 3D Gaussian represents the scene with a number of points, we use this icp loss to build the bridge of 3D Gaussian from different frame, which can further improve the scene representation and the camera tracking accuracy.

\noindent \textbf{Velocity Propagation and Silhouette Mask } We also conduct ablation study on the camera tracking in Tab.~\ref{tab:ablation2}: (1) the use of forward velocity propagation, (2) the use of a silhouette mask to mask out invalid map areas in the loss, and (3) setting the silhouette threshold to 0.99 instead of 0.5. The forward velocity propagation is vital for camera tracking. We can see that silhouette is also critical as without it tracking completely
fails.

\noindent \textbf{RGB and Depth Loss} Our system use both photometric (RGB) and depth loss in camera tracking and scene reconstruction.  In Tab.~\ref{tab:ablation2}, we ablate the decision to use both and investigate the performance of only use one or the other for both tracking and mapping. We conduct the ablation study on Replica dataset~\cite{replica} scene room0. With only depth, our method completely fails to track the camera trajectory, because the L1 depth loss doesn’t provide adequate information. Using
only an RGB loss successfully tracks the camera trajectory (although with more than 5x the error as using both). Both
the RGB and depth work together to achieve excellent results.

\section{Conclusion}
 In this paper, we revisit the existing GS-based SLAM systems and first demonstrate the geometric similarities of 3D Gaussian ellipsoids. Then, we propose a novel voxel-based compact GS-based SLAM system, reducing the number of redundant Gaussian ellipsoids without a decrease in performance. The proposed sliding window mask method and the geometry codebook improve the compactness of the scene representation, achieving faster training and rendering speed, and a significant reduction in memory usage. The proposed global bundle adjustment with icp loss further improves camera tracking accuracy and scene representation. The extensive experiments demonstrate that our work provides a comprehensive dense visual SLAM system, achieving high-fidelity performance, fast training, compactness, and real-time rendering. The experiments conducted on embedded platform shows the potential various real-world application in robotics and other embedded systems.

\noindent\textbf{Limitations}
In the current system, when handling fast motions, motion blur in the visual input leads to degraded performance, as both the RGB and depth signals become blurred. This negatively affects the tracking component, which relies on RGB and depth losses for optimization. On the other hand, in scenarios with a large number of dynamic objects, the system suffers from artifacts and drift. At present, our approach is still limited in effectively dealing with highly dynamic environments.
 \noindent\textbf{Acknowledgements} This work was supported by National Key R\&D Program of China (Grant No.2024YFB4708900).   It was also supported in part by the Natural Science Foundation of
China under Grant 62225309, U24A20278, 62361166632. This research is also supported by the National Research Foundation, Singapore, under the NRF Medium Sized Centre scheme (CARTIN), Maritime and Port Authority of Singapore under its Maritime Transformation Programme (Project No. SMI-2022-MTP-04), ASTAR under National Robotics Programme with Grant No. M22NBK0109. This work is supported by a Start-up Grant from Nanyang Technological University and jointly funded by the Singapore Ministry of Education (MOE) under a Tier-1 research grant. Corresponding author: Hesheng Wang (wanghesheng@sjtu.edu.cn)

 \noindent\textbf{Availability of data and materials} All the datasets used in the paper will be publicly available.

\newpage



\bibliography{main.bib}

@article{slam,
  title={Simultaneous localization and mapping: part I},
  author={Durrant-Whyte, Hugh and Bailey, Tim},
  journal={IEEE robotics \& automation magazine},
  volume={13},
  number={2},
  pages={99--110},
  year={2006},
  publisher={IEEE}
}

@InProceedings{imap,
    author    = {Sucar, Edgar and Liu, Shikun and Ortiz, Joseph and Davison, Andrew J.},
    title     = {iMAP: Implicit Mapping and Positioning in Real-Time},
    booktitle = {ICCV},
    month     = {October},
    year      = {2021},
    pages     = {6229-6238}
}

@inproceedings{NeRF,
author = {Mildenhall, Ben and Srinivasan, Pratul P. and Tancik, Matthew and Barron, Jonathan T. and Ramamoorthi, Ravi and Ng, Ren},
title = {NeRF: Representing Scenes as Neural Radiance Fields for View Synthesis},
booktitle = {ECCV},
year = {2020},
}

@InProceedings{niceslam,
    author    = {Zhu, Zihan and Peng, Songyou and Larsson, Viktor and Xu, Weiwei and Bao, Hujun and Cui, Zhaopeng and Oswald, Martin R. and Pollefeys, Marc},
    title     = {NICE-SLAM: Neural Implicit Scalable Encoding for SLAM},
    booktitle = {CVPR},
    month     = {June},
    year      = {2022},
    pages     = {12786-12796}
}

@article{replica,
  title={The Replica dataset: A digital replica of indoor spaces},
  author={Straub, Julian and Whelan, Thomas and Ma, Lingni and Chen, Yufan and Wijmans, Erik and Green, Simon and Engel, Jakob J and Mur-Artal, Raul and Ren, Carl and Verma, Shobhit and others},
  journal={arXiv preprint arXiv:1906.05797},
  year={2019}
}

@ARTICLE{orbslam,
  author={Mur-Artal, Raúl and Montiel, J. M. M. and Tardós, Juan D.},
  journal={IEEE Transactions on Robotics}, 
  title={ORB-SLAM: A Versatile and Accurate Monocular SLAM System}, 
  year={2015},
  volume={31},
  number={5},
  pages={1147-1163},
  doi={10.1109/TRO.2015.2463671}}

@ARTICLE{orbslam2,
  author={Mur-Artal, Raúl and Tardós, Juan D.},
  journal={IEEE Transactions on Robotics}, 
  title={ORB-SLAM2: An Open-Source SLAM System for Monocular, Stereo, and RGB-D Cameras}, 
  year={2017},
  volume={33},
  number={5},
  pages={1255-1262},
  doi={10.1109/TRO.2017.2705103}}

@Inproceedings{dtam,
  title={DTAM: Dense tracking and mapping in real-time},
  author={Newcombe, Richard A and Lovegrove, Steven J and Davison, Andrew J},
  booktitle={2011 international conference on computer vision},
  pages={2320--2327},
  year={2011},
  organization={IEEE}
}

@InProceedings{codeslam,
author = {Bloesch, Michael and Czarnowski, Jan and Clark, Ronald and Leutenegger, Stefan and Davison, Andrew J.},
title = {CodeSLAM — Learning a Compact, Optimisable Representation for Dense Visual SLAM},
booktitle = {Proceedings of the IEEE Conference on Computer Vision and Pattern Recognition (CVPR)},
month = {June},
year = {2018}
}

@inproceedings{kinectfusion,
  title={KinectFusion: real-time 3D reconstruction and interaction using a moving depth camera},
  author={Izadi, Shahram and Kim, David and Hilliges, Otmar and Molyneaux, David and Newcombe, Richard and Kohli, Pushmeet and Shotton, Jamie and Hodges, Steve and Freeman, Dustin and Davison, Andrew and others},
  booktitle={Proceedings of the 24th annual ACM symposium on User interface software and technology},
  pages={559--568},
  year={2011}
}

@InProceedings{scannet1,
author = {Dai, Angela and Chang, Angel X. and Savva, Manolis and Halber, Maciej and Funkhouser, Thomas and Niessner, Matthias},
title = {ScanNet: Richly-Annotated 3D Reconstructions of Indoor Scenes},
booktitle = {Proceedings of the IEEE Conference on Computer Vision and Pattern Recognition (CVPR)},
month = {July},
year = {2017}
}

@INPROCEEDINGS{tum,
  author={Sturm, Jürgen and Engelhard, Nikolas and Endres, Felix and Burgard, Wolfram and Cremers, Daniel},
  booktitle={2012 IEEE/RSJ International Conference on Intelligent Robots and Systems}, 
  title={A benchmark for the evaluation of RGB-D SLAM systems}, 
  year={2012},
  volume={},
  number={},
  pages={573-580},
  doi={10.1109/IROS.2012.6385773}}

@inproceedings{eslam,
  title={Eslam: Efficient dense slam system based on hybrid representation of signed distance fields},
  author={Johari, Mohammad Mahdi and Carta, Camilla and Fleuret, Fran{\c{c}}ois},
  booktitle={Proceedings of the IEEE/CVF Conference on Computer Vision and Pattern Recognition},
  pages={17408--17419},
  year={2023}
}

@inproceedings{coslam,
  title={Co-SLAM: Joint Coordinate and Sparse Parametric Encodings for Neural Real-Time SLAM},
  author={Wang, Hengyi and Wang, Jingwen and Agapito, Lourdes},
  booktitle={Proceedings of the IEEE/CVF Conference on Computer Vision and Pattern Recognition},
  pages={13293--13302},
  year={2023}
}

@INPROCEEDINGS{voxfusion,
  author={Yang, Xingrui and Li, Hai and Zhai, Hongjia and Ming, Yuhang and Liu, Yuqian and Zhang, Guofeng},
  booktitle={2022 IEEE International Symposium on Mixed and Augmented Reality (ISMAR)}, 
  title={Vox-Fusion: Dense Tracking and Mapping with Voxel-based Neural Implicit Representation}, 
  year={2022},
  volume={},
  number={},
  pages={499-507},
  doi={10.1109/ISMAR55827.2022.00066}}

@article{deng,
  title={Long-Term Visual Simultaneous Localization and Mapping: Using a Bayesian Persistence Filter-Based Global Map Prediction},
  author={Deng, Tianchen and Xie, Hongle and Wang, Jingchuan and Chen, Weidong},
  journal={IEEE Robotics \& Automation Magazine},
  volume={30},
  number={1},
  pages={36--49},
  year={2023},
  publisher={IEEE}
}

@article{xie,
  title={Robust Incremental Long-Term Visual Topological Localization in Changing Environments},
  author={Xie, Hongle and Deng, Tianchen and Wang, Jingchuan and Chen, Weidong},
  journal={IEEE Transactions on Instrumentation and Measurement},
  volume={72},
  pages={1--14},
  year={2022},
  publisher={IEEE}
}

@article{past,
  title={Past, present, and future of simultaneous localization and mapping: Toward the robust-perception age},
  author={Cadena, Cesar and Carlone, Luca and Carrillo, Henry and Latif, Yasir and Scaramuzza, Davide and Neira, Jos{\'e} and Reid, Ian and Leonard, John J},
  journal={IEEE Transactions on robotics},
  volume={32},
  number={6},
  pages={1309--1332},
  year={2016},
  publisher={IEEE}
}

@InProceedings{plgslam,
    author    = {Deng, Tianchen and Shen, Guole and Qin, Tong and Wang, Jianyu and Zhao, Wentao and Wang, Jingchuan and Wang, Danwei and Chen, Weidong},
    title     = {PLGSLAM: Progressive Neural Scene Represenation with Local to Global Bundle Adjustment},
    booktitle = {Proceedings of the IEEE/CVF Conference on Computer Vision and Pattern Recognition (CVPR)},
    month     = {June},
    year      = {2024},
    pages     = {19657-19666}
}

@inproceedings{splatam,
  title={Splatam: Splat track \& map 3d gaussians for dense rgb-d slam},
  author={Keetha, Nikhil and Karhade, Jay and Jatavallabhula, Krishna Murthy and Yang, Gengshan and Scherer, Sebastian and Ramanan, Deva and Luiten, Jonathon},
  booktitle={Proceedings of the IEEE/CVF Conference on Computer Vision and Pattern Recognition},
  pages={21357--21366},
  year={2024}
}

@inproceedings{gsslam,
  title={Gs-slam: Dense visual slam with 3d gaussian splatting},
  author={Yan, Chi and Qu, Delin and Xu, Dan and Zhao, Bin and Wang, Zhigang and Wang, Dong and Li, Xuelong},
  booktitle={Proceedings of the IEEE/CVF Conference on Computer Vision and Pattern Recognition},
  pages={19595--19604},
  year={2024}
}

@article{3dgs,
  title={3D Gaussian Splatting for Real-Time Radiance Field Rendering},
  author={Kerbl, Bernhard and Kopanas, Georgios and Leimk{\"u}hler, Thomas and Drettakis, George},
  journal={ACM Transactions on Graphics},
  volume={42},
  number={4},
  year={2023}
}

@inproceedings{monogs,
  title={Gaussian splatting slam},
  author={Matsuki, Hidenobu and Murai, Riku and Kelly, Paul HJ and Davison, Andrew J},
  booktitle={Proceedings of the IEEE/CVF Conference on Computer Vision and Pattern Recognition},
  pages={18039--18048},
  year={2024}
}

@article{gaussianslam,
  title={Gaussian-SLAM: Photo-realistic Dense SLAM with Gaussian Splatting},
  author={Yugay, Vladimir and Li, Yue and Gevers, Theo and Oswald, Martin R},
  journal={arXiv preprint arXiv:2312.10070},
  year={2023}
}

@article{droidslam,
  title={Droid-slam: Deep visual slam for monocular, stereo, and rgb-d cameras},
  author={Teed, Zachary and Deng, Jia},
  journal={Advances in neural information processing systems},
  volume={34},
  pages={16558--16569},
  year={2021}
}

@inproceedings{pointslam,
  title={Point-slam: Dense neural point cloud-based slam},
  author={Sandstr{\"o}m, Erik and Li, Yue and Van Gool, Luc and Oswald, Martin R},
  booktitle={Proceedings of the IEEE/CVF International Conference on Computer Vision},
  pages={18433--18444},
  year={2023}
}

@inproceedings{compactgs1,
  title={Compact 3d gaussian representation for radiance field},
  author={Lee, Joo Chan and Rho, Daniel and Sun, Xiangyu and Ko, Jong Hwan and Park, Eunbyung},
  booktitle={Proceedings of the IEEE/CVF Conference on Computer Vision and Pattern Recognition},
  pages={21719--21728},
  year={2024}
}

@article{lightgaussian,
  title={Lightgaussian: Unbounded 3d gaussian compression with 15x reduction and 200+ fps},
  author={Fan, Zhiwen and Wang, Kevin and Wen, Kairun and Zhu, Zehao and Xu, Dejia and Wang, Zhangyang and others},
  journal={Advances in neural information processing systems},
  volume={37},
  pages={140138--140158},
  year={2024}
}

@inproceedings{compactgs2,
  title={Compact 3d scene representation via self-organizing gaussian grids},
  author={Morgenstern, Wieland and Barthel, Florian and Hilsmann, Anna and Eisert, Peter},
  booktitle={European Conference on Computer Vision},
  pages={18--34},
  year={2024},
  organization={Springer}
}

@article{gradient,
  title={Estimating or propagating gradients through stochastic neurons for conditional computation},
  author={Bengio, Yoshua and L{\'e}onard, Nicholas and Courville, Aaron},
  journal={arXiv preprint arXiv:1308.3432},
  year={2013}
}

@article{soundstream,
  title={Soundstream: An end-to-end neural audio codec},
  author={Zeghidour, Neil and Luebs, Alejandro and Omran, Ahmed and Skoglund, Jan and Tagliasacchi, Marco},
  journal={IEEE/ACM Transactions on Audio, Speech, and Language Processing},
  volume={30},
  pages={495--507},
  year={2021},
  publisher={IEEE}
}

@article{Kintinuous,
  title={Real-time large-scale dense RGB-D SLAM with volumetric fusion},
  author={Whelan, Thomas and Kaess, Michael and Johannsson, Hordur and Fallon, Maurice and Leonard, John J and McDonald, John},
  journal={The International Journal of Robotics Research},
  volume={34},
  number={4-5},
  pages={598--626},
  year={2015},
  publisher={Sage Publications Sage UK: London, England}
}

@article{elasticfusion,
  title={ElasticFusion: Real-time dense SLAM and light source estimation},
  author={Whelan, Thomas and Salas-Moreno, Renato F and Glocker, Ben and Davison, Andrew J and Leutenegger, Stefan},
  journal={The International Journal of Robotics Research},
  volume={35},
  number={14},
  pages={1697--1716},
  year={2016},
  publisher={SAGE Publications Sage UK: London, England}
}

@inproceedings{raft,
  title={Raft: Recurrent all-pairs field transforms for optical flow},
  author={Teed, Zachary and Deng, Jia},
  booktitle={Computer Vision--ECCV 2020: 16th European Conference, Glasgow, UK, August 23--28, 2020, Proceedings, Part II 16},
  pages={402--419},
  year={2020},
  organization={Springer}
}

@article{dpvo,
  title={Deep patch visual odometry},
  author={Teed, Zachary and Lipson, Lahav and Deng, Jia},
  journal={Advances in Neural Information Processing Systems},
  volume={36},
  year={2024}
}

@inproceedings{rtg-slam,
  title={Rtg-slam: Real-time 3d reconstruction at scale using gaussian splatting},
  author={Peng, Zhexi and Shao, Tianjia and Liu, Yong and Zhou, Jingke and Yang, Yin and Wang, Jingdong and Zhou, Kun},
  booktitle={ACM SIGGRAPH 2024 Conference Papers},
  pages={1--11},
  year={2024}
}

@inproceedings{scaffoldgs,
  title={Scaffold-gs: Structured 3d gaussians for view-adaptive rendering},
  author={Lu, Tao and Yu, Mulin and Xu, Linning and Xiangli, Yuanbo and Wang, Limin and Lin, Dahua and Dai, Bo},
  booktitle={Proceedings of the IEEE/CVF Conference on Computer Vision and Pattern Recognition},
  pages={20654--20664},
  year={2024}
}

@inproceedings{sgsslam,
  title={Sgs-slam: Semantic gaussian splatting for neural dense slam},
  author={Li, Mingrui and Liu, Shuhong and Zhou, Heng and Zhu, Guohao and Cheng, Na and Deng, Tianchen and Wang, Hongyu},
  booktitle={European Conference on Computer Vision},
  pages={163--179},
  year={2024},
  organization={Springer}
}

@inproceedings{loopsplat,
  title={Loopsplat: Loop closure by registering 3d gaussian splats},
  author={Zhu, Liyuan and Li, Yue and Sandstr{\"o}m, Erik and Huang, Shengyu and Schindler, Konrad and Armeni, Iro},
  booktitle={2025 International Conference on 3D Vision (3DV)},
  pages={156--167},
  year={2025},
  organization={IEEE}
}

@inproceedings{photoslam,
  title={Photo-slam: Real-time simultaneous localization and photorealistic mapping for monocular stereo and rgb-d cameras},
  author={Huang, Huajian and Li, Longwei and Cheng, Hui and Yeung, Sai-Kit},
  booktitle={Proceedings of the IEEE/CVF Conference on Computer Vision and Pattern Recognition},
  pages={21584--21593},
  year={2024}
}

@article{mcnslam,
      title={MCN-SLAM: Multi-Agent Collaborative Neural SLAM with Hybrid Implicit Neural Scene Representation}, 
      author={Tianchen Deng and Guole Shen and Xun Chen and Shenghai Yuan and Hongming Shen and Guohao Peng and Zhenyu Wu and Jingchuan Wang and Lihua Xie and Danwei Wang and Hesheng Wang and Weidong Chen},
      journal={arXiv preprint arXiv:2506.18678},
      year={2025},
}

@inproceedings{mneslam,
  title={Mne-slam: Multi-agent neural slam for mobile robots},
  author={Deng, Tianchen and Shen, Guole and Xun, Chen and Yuan, Shenghai and Jin, Tongxin and Shen, Hongming and Wang, Yanbo and Wang, Jingchuan and Wang, Hesheng and Wang, Danwei and others},
  booktitle={Proceedings of the Computer Vision and Pattern Recognition Conference},
  pages={1485--1494},
  year={2025}
}

@article{neslam,
  title={Neslam: Neural implicit mapping and self-supervised feature tracking with depth completion and denoising},
  author={Deng, Tianchen and Wang, Yanbo and Xie, Hongle and Wang, Hesheng and Guo, Rui and Wang, Jingchuan and Wang, Danwei and Chen, Weidong},
  journal={IEEE Transactions on Automation Science and Engineering},
  year={2025},
  publisher={IEEE}
}

@ARTICLE{structureslam,
  author={Liu, Shuhong and Deng, Tianchen and Zhou, Heng and Li, Liuzhuozheng and Wang, Hongyu and Wang, Danwei and Li, Mingrui},
  journal={IEEE Transactions on Automation Science and Engineering}, 
  title={MG-SLAM: Structure Gaussian Splatting SLAM With Manhattan World Hypothesis}, 
  year={2025},
  volume={22},
  number={},
  pages={17034-17049},
  doi={10.1109/TASE.2025.3575772}}

@inproceedings{loopyslam,
  title={Loopy-slam: Dense neural slam with loop closures},
  author={Liso, Lorenzo and Sandstr{\"o}m, Erik and Yugay, Vladimir and Van Gool, Luc and Oswald, Martin R},
  booktitle={Proceedings of the IEEE/CVF Conference on Computer Vision and Pattern Recognition},
  pages={20363--20373},
  year={2024}
}

@inproceedings{goslam,
  title={Go-slam: Global optimization for consistent 3d instant reconstruction},
  author={Zhang, Youmin and Tosi, Fabio and Mattoccia, Stefano and Poggi, Matteo},
  booktitle={Proceedings of the IEEE/CVF International Conference on Computer Vision},
  pages={3727--3737},
  year={2023}
}

@article{fmgs,
  title={Fmgs: Foundation model embedded 3d gaussian splatting for holistic 3d scene understanding},
  author={Zuo, Xingxing and Samangouei, Pouya and Zhou, Yunwen and Di, Yan and Li, Mingyang},
  journal={International Journal of Computer Vision},
  pages={1--17},
  year={2024},
  publisher={Springer}
}

@article{slam2,
  title={A survey on global lidar localization: Challenges, advances and open problems},
  author={Yin, Huan and Xu, Xuecheng and Lu, Sha and Chen, Xieyuanli and Xiong, Rong and Shen, Shaojie and Stachniss, Cyrill and Wang, Yue},
  journal={International Journal of Computer Vision},
  volume={132},
  number={8},
  pages={3139--3171},
  year={2024},
  publisher={Springer}
}

@article{pseudo,
  title={Pseudo-Plane Regularized Signed Distance Field for Neural Indoor Scene Reconstruction},
  author={Li, Jing and Yu, Jinpeng and Wang, Ruoyu and Gao, Shenghua},
  journal={International Journal of Computer Vision},
  pages={1--19},
  year={2024},
  publisher={Springer}
}

@article{hscnet++,
  title={Hscnet++: Hierarchical scene coordinate classification and regression for visual localization with transformer},
  author={Wang, Shuzhe and Laskar, Zakaria and Melekhov, Iaroslav and Li, Xiaotian and Zhao, Yi and Tolias, Giorgos and Kannala, Juho},
  journal={International Journal of Computer Vision},
  volume={132},
  number={7},
  pages={2530--2550},
  year={2024},
  publisher={Springer}
}

@article{slam3,
  title={4seasons: Benchmarking visual slam and long-term localization for autonomous driving in challenging conditions},
  author={Wenzel, Patrick and Yang, Nan and Wang, Rui and Zeller, Niclas and Cremers, Daniel},
  journal={International Journal of Computer Vision},
  pages={1--23},
  year={2024},
  publisher={Springer}
}

@inproceedings{snislam,
  title={Sni-slam: Semantic neural implicit slam},
  author={Zhu, Siting and Wang, Guangming and Blum, Hermann and Liu, Jiuming and Song, Liang and Pollefeys, Marc and Wang, Hesheng},
  booktitle={Proceedings of the IEEE/CVF Conference on Computer Vision and Pattern Recognition},
  pages={21167--21177},
  year={2024}
}

@article{semgauss,
  title={Semgauss-slam: Dense semantic gaussian splatting slam},
  author={Zhu, Siting and Qin, Renjie and Wang, Guangming and Liu, Jiuming and Wang, Hesheng},
  journal={arXiv preprint arXiv:2403.07494},
  year={2024}
}

@ARTICLE{wz1,
  author={Wang, Zhong and Zhang, Lin and Wang, Hesheng},
  journal={IEEE Transactions on Circuits and Systems for Video Technology}, 
  title={S²KAN-SLAM: Elastic Neural LiDAR SLAM With SDF Submaps and Kolmogorov-Arnold Networks}, 
  year={2025},
  volume={35},
  number={8},
  pages={7618-7630},
  doi={10.1109/TCSVT.2025.3550871}}

@ARTICLE{wz2,
  author={Wang, Zhong and Zhang, Lin and Shen, Ying and Zhou, Yicong},
  journal={IEEE Transactions on Multimedia}, 
  title={D-LIOM: Tightly-Coupled Direct LiDAR-Inertial Odometry and Mapping}, 
  year={2023},
  volume={25},
  number={},
  pages={3905-3920},
  doi={10.1109/TMM.2022.3168423}}

@article{deng2025best3dscenerepresentation,
      title={What Is The Best 3D Scene Representation for Robotics? From Geometric to Foundation Models}, 
      author={Tianchen Deng and Yue Pan and Shenghai Yuan and Dong Li and Chen Wang and Mingrui Li and Long Chen and Lihua Xie and Danwei Wang and Jingchuan Wang and Javier Civera and Hesheng Wang and Weidong Chen},
      year={2025},
      journal={arXiv preprint arXiv:2512.03422}, 
}
\end{document}